%% file: main.tex
\documentclass{article}
\usepackage{etoc}
\usepackage[preprint]{neurips_2026}
\usepackage[utf8]{inputenc}
\usepackage[T1]{fontenc}
\usepackage{hyperref}
\usepackage{url}
\usepackage{booktabs}
\usepackage{amsfonts}
\usepackage{amsmath}
\usepackage{nicefrac}
\usepackage{microtype}
\usepackage{xcolor}
\usepackage{graphicx}
\usepackage{algorithm}
\usepackage{natbib}
\usepackage{algorithmic}

\usepackage{tikz}
\usetikzlibrary{positioning, fit, backgrounds, arrows.meta, calc}

\newcommand\TODO[1]{\textcolor{red}{#1}}

\title{
AlphaExploitem:
Going Beyond the Nash Equilibrium in Poker
by Learning to Exploit Suboptimal Play
}

\author{%
    Vlad Murgoci \\
    Delft University of Technology \\
    2628 CD Delft, The Netherlands \\
    \texttt{v.murgoci@tudelft.nl}
    \AND
    Matthijs Spaan \\
    Department of Intelligent Systems \\
    Delft University of Technology \\
    2628 CD Delft, The Netherlands \\
    \texttt{M.T.J.Spaan@tudelft.nl}
    \AND
    Yaniv Oren \\
    Department of Intelligent Systems\\
    Delft University of Technology\\
    2628 CD Delft, The Netherlands \\
    \texttt{y.oren@tudelft.nl} \\
}

\begin{document}
\etocdepthtag.toc{main}
\maketitle

\input{Sections/Abstract}


\input{Sections/Introduction}


\input{Sections/Background}


\input{Sections/Prior-Work}

\input{Sections/Contributions}


\input{Sections/Results}

\input{Sections/Conclusion}


{
\small
\bibliographystyle{plainnat}
\bibliography{references}
}


\appendix
\etocdepthtag.toc{appendix}

\newpage

{
  \etocsettagdepth{main}{none}
  \etocsettagdepth{appendix}{subsection}
  \etocsettocstyle{\section*{Appendix Contents}}{}
  \tableofcontents
}

\newpage

\input{Sections/Appendices/Limitations}

\input{Sections/Appendices/Player-Types}

\input{Sections/Appendices/ToyStrategies}

\input{Sections/Appendices/NashEquilibria}

\input{Sections/Appendices/BestResponseCeilings}

\input{Sections/Appendices/Hyperparameters}

\newpage
\input{checklist.tex}

\end{document}

%% file: Sections/Abstract.tex
\begin{abstract}

Poker is an imperfect information game that has served as a long-standing benchmark for decision-making under uncertainty. 
To maximize utility beyond the Nash equilibrium, an agent can deviate from Nash-equilibrium policies to exploit suboptimal play. 
We introduce \textbf{AlphaExploitem}, which extends the competitive RL poker agent AlphaHoldem by using a hierarchical transformer encoder that enables reasoning over previously played hands and modifying the training procedure with the inclusion of  a diverse pool of exploitable opponents to facilitate learning to exploit.
We train and evaluate AlphaExploitem on two standard benchmarks for imperfect-information games.
Empirically, AlphaExploitem successfully exploits weak play by \textit{both in- and out-of-distribution opponents}, without losing performance against NE opponents.
\end{abstract}

%

%% file: Sections/Introduction.tex
\section{Introduction}
\label{sec:introduction}

Poker is a family of card games whose incomplete information setting, stochastic elements, and deep strategic complexity have made it a long-standing challenge for artificial intelligence~\citep{ComputerPoker}.
Unlike perfect information games such as chess and Go, poker requires agents to reason under uncertainty about hidden cards while navigating a vast strategic decision space spanning multiple betting rounds~\citep{Pluribus}.

State-of-the-art poker agents approximate the Nash Equilibrium (NE): strategies from which no player can improve by unilateral deviation~\citep{NashEq}.
Systems such as DeepStack~\citep{DeepStack}, Libratus~\citep{Libratus}, Pluribus~\citep{Pluribus}, and AlphaHoldem~\citep{AlphaHoldem} have achieved high performance in various poker formats by learning low-exploitability, approximately-NE strategies.
However, while NE guarantees a non-negative expected return against any opponent, it does not maximize returns against \textit{out-of-equilibrium} play.
Against suboptimal opponents, substantial additional expected value (EV) can be captured by deviating from the equilibrium to exploit observed flaws in the opponent's strategy.

Competitive poker is rarely played as a single \textit{hand}, however.
A hand is one full deal of cards from shuffling the deck to determining the winner of the wagered money, i.e., a single phase of the entire game. 
Sessions may consist of hundreds to thousands of hands between the same opponents.
This creates a repeated-interaction structure that proficient players often leverage \citep{HumansExploit}, adapting their strategy in response to patterns observed in an opponent's play. 
However, state-of-the-art NE-approximating agents treat every hand as an independent encounter: their decisions depend only on the current hand's cards and betting sequence, discarding any additional information available from the session so far.

Prior work on exploiting suboptimal poker opponents has largely relied on hand-crafted summary features of opponent behavior: conditional strategies based on aggregated game statistics \citep{Loki, Poki, Explicit-opponent-model}, recurrent networks consuming pre-computed summary features \citep{lstm-exploiter1, lstm-exploiter2, rnn-exploiter}. \citet{ICE} train a causal transformer via algorithm distillation \citep{AlgorithmDistillation}. Although similar in architecture, the method has been criticized for its training technique that does not guarantee or evaluate on human and rational flawed strategies.

In this work, we introduce \textbf{AlphaExploitem}, an actor-critic agent trained through Proximal Policy Optimization (PPO) \citep{PPO} for which trajectories are generated via \textit{K-best-league} self-play \citep{AlphaHoldem}, a \textit{long-tail buffer} of past iterations of the main agent, and a set of hand-crafted exploitable opponents.
To exploit weak play, AlphaExploitem extends AlphaHoldem's \citep{AlphaHoldem} architecture with a hierarchical transformer encoder that processes the raw tokenized history of previously played hands within a session.
By using separate learned embedding tables for different types of observations, we provide complete tokenized information over the opponent's behavior and the environment's dynamics, delegating the intrinsic task of processing game data to the model itself rather than to human-built feature aggregators.

We evaluate on two standard imperfect-information benchmarks, Kuhn Poker \citep{kuhn2016simplified} and Leduc Hold'em \citep{Leduc}.
AlphaExploitem captures more than twice the per-hand EV of the AlphaHoldem-style league-only baseline, while remaining close to the NE and roughly twice the EV of the same trained policy without access to hand histories.
In addition, we observe that AlphaExploitem generalizes effectively to unseen exploitable opponents, exploiting out-of-distribution agents as well as in distribution.
A reproducible evaluation ensemble is also provided in Appendix \ref{app:toy-strategies}, and to the best of our knowledge, this is the first work that provides means of comparison for future research on exploiting suboptimal opponents in poker.

Through this work, we hope to further research and comparison benchmarks on exploitative play in poker, and more broadly on learning how to leverage complete observable information over weak playstyles in repeated interactions in order to maximize returns.
Since poker is mainly played between human players, our research can be used as a foundation for building tools that can help players find weaknesses in their game and as well as in their opponents' strategies, and to train towards effectively adapting.

%

%% file: Sections/Background.tex
\section{Background}
\label{sec:background}

\textbf{Poker} is a card game which usually involves \textit{hidden} private cards, sometimes \textit{shared} community cards, and rounds of \textit{betting} with chips (in-game money).
A player wins either by having the best hand at \textit{showdown} or by making all others fold (give up) their hand.
Because cards are hidden, \textit{bluffing} is an important strategic element.
Poker is usually formalized through two complementary lenses: as a \textit{partially observable decision problem} from a single player's perspective, and as a multi-agent strategic interaction problem.

We refer to the player we're interested in as the \textit{hero} while poker convention denotes the other competing players as \textit{villains}.
Each hand begins with two forced pre-deal wagers, the \textit{small blind} (SB) and the \textit{big blind} (BB), which seed the central wager pool --- the \textit{pot}.
The main utility metric is \textit{big blinds per hand} (BBs/hand) where one big blind is the standard unit of wager.
At each decision the active player chooses among four canonical actions: \textit{fold} (forfeit the hand), \textit{check} (decline to wager when no bet is pending), \textit{call} (match a pending bet), and \textit{bet}/\textit{raise} (commit chips beyond what is already wagered).
The hand ends either when one player folds --- at which point the other wins the pot uncontested --- or at showdown, when the surviving players reveal their \textit{hole cards} and the highest-ranked hand wins.
Players are commonly described in terms of stylistic \textit{archetypes} that summarize their tendencies across many hands.
We use these archetypes both as training opponents and as a mental model for the kind of cross-hand information our agent is meant to exploit.
More details about player archetypes can be found in Appendix~\ref{app:players}

\textbf{Imperfect information games.}
We adopt standard extensive-form game notation: each player $i$ has a behavioral strategy $\sigma_i$ mapping their \textit{information sets} --- decision points the player cannot distinguish --- to distributions over legal actions, with terminal utility $u_i$, and a strategy profile $\sigma = (\sigma_i)_{i \in N}$ collects all players' policies.
Heads-up poker is the two-player zero-sum special case ($u_1 = -u_2$), which makes the equilibrium and exploitability concepts of the next paragraph well-defined and, for Kuhn and Leduc poker, exactly computable.

In poker, each information set $I \in \mathcal{I}_i$ corresponds to a decision node at which the acting player knows the public game history and their own private cards but cannot distinguish between histories that differ only in the opponent's private cards.
Heads-up poker instantiates the two-player zero-sum setting ($N = \{1, 2\}$, $u_1(z) = -u_2(z)$), which makes equilibrium and exploitability --- introduced next --- well-defined and finite-dimensionally computable for the environments we study.

\textbf{Nash Equilibrium and Exploitability.}
A Nash Equilibrium~\citep{NashEq} is a strategy profile~$\sigma^*$ from which no player can improve their expected payoff by unilateral deviation: $u_i(\sigma^*) \ge \max_{\sigma_i'} u_i(\sigma_i', \sigma^*_{-i})$ for all $i$.
Exploitability~\citep{Exploitability} measures how much an opponent can gain by deviating against a given strategy to maximize utility.
NE has zero exploitability.
While NE strategies are safe, they sacrifice the additional utility available against opponents who are not at the equilibrium.

\textbf{Modeling Poker as a partially observable Markov Decision Process}
Heads-up (two-player) poker is naturally a \emph{partially observable stochastic game}: both players see the public betting sequence and any community cards, but each player's hole cards are hidden from the other.

Fixing a stationary opponent policy reduces this to a single-agent \emph{partially observable Markov decision process} for the hero, with two consequences for the rest of the paper.
The resulting transition and observation distributions are \emph{opponent-specific}, so an agent that conditions on observed opponent patterns is effectively learning a per-opponent transition model.

\textbf{AlphaHoldem.}
AlphaHoldem~\citep{AlphaHoldem} is a competitive end-to-end RL poker agent that reaches CFR-level play~\citep{DeepStack} via two design choices we inherit: architecture and training technique. 
The authors employ self-play training with PPO through a \textit{K-best league}, a fixed-size pool of past checkpoints where past iterations that achieve higher ELOs survive in the league, while low-performing agents are eliminated.
AlphaHoldem uses a pseudo-siamese architecture: two separate networks process information about cards and player actions of the current poker hand reflecting the fact that these observations capture different aspects of the game. The output of the two networks is fused only at the policy and value heads.

We choose AlphaHoldem as our foundation because it is a competitive end-to-end RL agent that reaches CFR-level play at a fraction of the compute.
Crucially, its architecture cleanly separates the card and action streams and fuses them only at the policy and value heads, leaving a natural extension point for a cross-hand module as a third input stream.

%

%% file: Sections/Prior-Work.tex
\section{Related Work}
\label{sec:prior-work}

\textbf{RL for Poker.} Other additional techniques complementing the approaches discussed in Section~\ref{sec:introduction} used for NE play in Poker are: Neural Fictitious Self-Play (NFSP)~\citep{NFSP} which combines DQN-based best-response learning with supervised average-strategy regression to approach Nash equilibrium, Deep CFR~\citep{DeepCFR} and its model-free successor DREAM~\citep{DREAM} replace tabular regret with deep networks, scaling regret minimization beyond hand-crafted abstractions while ReBeL~\citep{ReBeL} further couples deep RL with search to attain strong play in large imperfect-information games.
These systems aim for low exploitability rather than opponent-specific adaptation.

\textbf{Exploiting suboptimal opponents.}
Work on exploiting suboptimal poker opponents can be broadly grouped by how opponent behavior is represented.
Early approaches relied on hand-crafted statistics accumulated over the course of play and passed them to conditional decision-making modules, including expert systems and feed-forward networks \citep{Loki, Poki, KuhnExploiter, Explicit-opponent-model}.
More recent work replaced these manually designed pipelines with learned sequence models, such as recurrent architectures, but still operated on per-hand summaries or cumulative behavioral statistics rather than raw interaction histories \citep{lstm-exploiter1, lstm-exploiter2, rnn-exploiter}. 
In contrast, \citep{ICE} processes raw hand histories directly with a causal transformer, avoiding explicit feature engineering.
However, its training and evaluation protocol has been questioned for not clearly ensuring exposure to a sufficiently diverse set of weak opponents.

\textbf{LLM-based approaches.} PokerGPT~\citep{pokergpt} leverages large language models by casting the poker state as natural language prompts, a methodologically distinct direction from the specialized numerical encoders used by our work and by the AlphaHoldem family. 
Current results indicate that LLMs struggle to match the performance of dedicated architectures especially against NE \citep{LLMBenchmark}.

%% file: Sections/Contributions.tex
\section{AlphaExploitem - Transformer-based Exploitation}                                                                                              
\label{sec:method}                                

AlphaExploitem extends AlphaHoldem to enable cross-hand opponent-specific adaptation without abandoning the inductive biases that make the base architecture compute-efficient.
The design follows three ideas.
First, we \textbf{preserve AlphaHoldem's architecture} for current-hand information in order to preserve the already proven efficiency over singular hand decisions. 
Second, we \textbf{separate responsibilities between streams}: the transformer sees only tokens from previously completed hands, while decisions within the current hand remain the responsibility of the original modules, thereby following the principle of dividing different types of observations to different network modules.
These principles yield an architecture in which the benefit of cross-hand context can be isolated (by fully masking the transformer input during inference).
Third, we \textbf{enhance} the existing K-best league training technique with training against a fixed set of hand-crafted flawed policies, and a dynamic collection of past checkpoints of the agent itself, to ensure that the agent is exposed to a wide variety of exploitable patterns during training.

\subsection{Encoding Poker Games}
\label{subsec:data}

For each decision, AlphaExploitem assembles three input tensors from the running game state: the cards available to the agent, the action sequence of the current hand, and a data structure containing all previously completed hands against the current opponent.

\textbf{Cards and Current-hand actions}
We replicate AlphaHoldem's encoding of card information and action information as separate one-hot encoded arrays.

\textbf{Cross-hand history.}
The history stream is a sequence of tokens recording every observable event from previously completed hands: hero actions, opponent actions, and revealed hole and community cards.
Each token carries a type tag (agent action, opponent action, private card, community card, opponent card), so that the encoder can route different token types through separate embedding tables.
We choose to exclude reward signals from the history stream as they can be inferred from the sequence of actions and cards.
In contrast to prior work, we do not aggregate hand information.

\subsection{Architecture}
\label{subsec:architecture}

Figure~\ref{fig:arch} summarizes the overall architecture.
\input{figures/Architecture/architecture}

\textbf{Current-hand encoders.}
The card and action tensors are each projected through a fully-connected layer.
We retain AlphaHoldem's separation of card and action processing to preserve representational separation.

\textbf{Hierarchical history transformer.}
We encode the history stream with a hierarchical transformer~\citep{Transformers} that mirrors poker's natural temporal granularity.
A \textbf{within-hand encoder} is applied independently to each past hand: it embeds the hand's tokens via type-specific embedding tables, runs self-attention over them, and pools the result to a single summary vector $h_i$.
An \textbf{across-hand encoder} then processes the resulting sequence of summaries $\{h_1, \ldots, h_M\}$ to produce a session-level context vector $z$.
In contrast to prior work, we delegate the task of summarizing hand information to the model itself rather than relying on hand-crafted statistics.

\textbf{Current-hand exclusion from the transformer.}
The transformer is restricted to tokens from hands already completed.
The current hand's card and action information is handled exclusively by the current-hand encoders.
This separation serves two purposes.
First, it produces a structural decomposition --- the transformer attends to \textit{cross-hand} information while the dedicated card and action encoders handle \textit{in-hand} reasoning.
Second, it prevents the transformer from becoming the primary decision-maker within a single hand.

\textbf{Fusion and output heads.}
The three encoded streams are concatenated, layer-normalized, and passed through a shared hidden network.
The representation then splits into an actor head producing an action distribution via softmax and a critic head producing a scalar value estimate.

\subsection{Training}
\label{subsec:training}

During each training epoch, we generate trajectories of experience from 3 opponent pools: the league, the flawed toy policies, and the dynamic buffer of past versions of the agent.
To generate training minibatches, we uniformly sample from the combined pool of trajectories and apply PPO updates to the agent's parameters.

\textbf{Adapted K-best league self-play.}
Based on AlphaHoldem~\citep{AlphaHoldem}, we train via self-play against opponents sampled from a fixed-size league pool.
The league maintains ELO ratings~\citep{Elo1978TheRO} updated based on head-to-head match outcomes, weighted by the winning margin.
At checkpoint intervals, the current agent competes against all league members.
If its ELO exceeds the lowest-rated member, it replaces that member.
This framework provides a diverse and progressively stronger set of training opponents and drives convergence toward a robust baseline policy.

\textbf{Exposure to flawed policies.}
In addition to the league, we add trajectories generated against a fixed ensemble of hand-crafted weak policies that cover common human-like archetypes.
The set of training toys has been crafted with domain expertise to cover a range of exploitable patterns, such as over-aggression or over-conservatism.
We take the standard assumption that flawed opponents do not adapt to the agent's behavior, following a stationary policy.
While this training setup has been used before \citep{L2E}, its combination with a transformer conditioned on session-level history is what enables the agent to learn \textit{adaptive} exploitation.

\textbf{Long-tail buffer self-play.}
In order to enable the agent to generalize to diverse policies, we use a long-tail buffer of previous checkpoints of the agent.
Unlike the toy set, where opponents have stationary policies, the snapshot opponent is a frozen past version of the main agent that does have access to its own cross-hand history channel.
As the agent cycles through different strategy profiles throughout training, playing against past, possibly flawed iterations exposes the main agent to various play patterns that are not contained within the set of hand-crafted policies.
The buffer acts as a bridge between the toy set and the league, providing a dynamic source of diverse opponents that possibly present various exploitable patterns and counter-adaptive pressure without the risk of overfitting.

%

%% file: figures/Architecture/architecture.tex
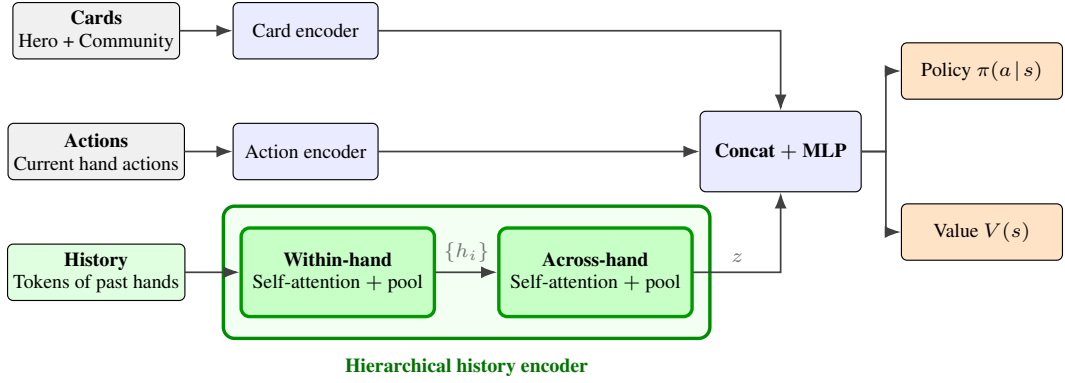
\begin{figure}[t]
\centering
\resizebox{\linewidth}{!}{%
\begin{tikzpicture}[
    font=\footnotesize,
    >=Latex,
    every node/.style={align=center},
    inp/.style={draw, rounded corners=2pt, fill=gray!12,
                minimum height=7mm, minimum width=20mm, inner sep=2pt,
                font=\scriptsize},
    inph/.style={draw, rounded corners=2pt, fill=green!12,
                minimum height=7mm, minimum width=20mm, inner sep=2pt,
                font=\scriptsize},
    enc/.style={draw, rounded corners=2pt, fill=blue!8,
                minimum height=7mm, minimum width=18mm, inner sep=2pt,
                font=\scriptsize},
    txblk/.style={draw=green!60!black, very thick, rounded corners=3pt,
                fill=green!22,
                minimum height=11mm, minimum width=24mm, inner sep=3pt,
                font=\scriptsize},
    fuse/.style={draw, rounded corners=2pt, fill=blue!8,
                minimum height=10mm, minimum width=20mm, inner sep=3pt,
                font=\scriptsize},
    head/.style={draw, rounded corners=2pt, fill=orange!22,
                minimum height=7mm, minimum width=20mm, inner sep=2pt,
                font=\scriptsize},
    grp/.style={draw=green!55!black, very thick, rounded corners=4pt,
                inner xsep=2mm, inner ysep=2.5mm,
                fill=green!5},
    arr/.style={->, semithick, draw=black!75, line cap=round},
    lbl/.style={font=\scriptsize\itshape, text=black!55},
]

\def\yc{ 1.5}  
\def\ya{ 0.0}  
\def\yh{-1.5}  

\node[inp]  (cards) at (0, \yc) {\textbf{Cards}\\\scriptsize Hero + Community};
\node[inp]  (acts)  at (0, \ya) {\textbf{Actions}\\\scriptsize Current hand actions};
\node[inph] (hist)  at (0, \yh) {\textbf{History}\\\scriptsize Tokens of past hands};

\node[enc] (cardEnc) at (2.6, \yc) {Card encoder};
\node[enc] (actEnc)  at (2.6, \ya) {Action encoder};

\node[txblk] (within) at (3.0, \yh) {\textbf{Within-hand}\\\scriptsize Self-attention $+$ pool};
\node[txblk] (across) at (6.2, \yh) {\textbf{Across-hand}\\\scriptsize Self-attention $+$ pool};

\begin{scope}[on background layer]
  \node[grp, fit=(within)(across)] (grpTx) {};
\end{scope}
\node[font=\scriptsize\bfseries, text=green!45!black,
      anchor=north] at ($(grpTx.south)+(0,-1mm)$) {Hierarchical history encoder};

\node[fuse] (mlp)  at (8.5, 0)    {\textbf{Concat $+$ MLP}};
\node[head] (pi)   at (11.0, 1.0) {Policy $\pi(a\!\mid\!s)$};
\node[head] (V)    at (11.0,-1.0) {Value $V(s)$};

\draw[arr] (cards)  -- (cardEnc);
\draw[arr] (acts)   -- (actEnc);
\draw[arr] (hist)   -- (within);
\draw[arr] (within) -- node[lbl, above]{$\{h_i\}$} (across);

\draw[arr] (cardEnc.east) -| (mlp.north);
\draw[arr] (actEnc.east)  -- (mlp.west);
\draw[arr] (across.east)  -| node[lbl, above, pos=0.25]{$z$} (mlp.south);

\draw[arr] (mlp.east) -- ++(0.3,0) |- (pi.west);
\draw[arr] (mlp.east) -- ++(0.3,0) |- (V.west);

\end{tikzpicture}%
}
\caption{\textbf{AlphaExploitem architecture (left $\to$ right).}
Three input streams---the cards available to the hero, the current hand's action history, and a tokenized record of all past hands against the same opponent---are fused via a shared MLP that splits into policy and value heads.
Our architectural contribution is the hierarchical history encoder (highlighted): a within-hand transformer summarizes each completed past hand to a vector $h_i$, and an across-hand transformer attends over the per-hand summaries to produce a session-level context $z$.}
\label{fig:arch}
\end{figure}

%% file: Sections/Results.tex
\section{Experiments}
\label{sec:results}

AlphaExploitem is evaluated on Leduc and Kuhn Poker against diverse opponent archetypes. 
Our results support four claims: 
(i) the agent exploits in-distribution suboptimal opponents, with gains attributable to cross-hand context; 
(ii) the exploitation demonstrates generalization to out-of-distribution opponents; 
(iii) the agent identifies strong policies and play accordingly against them;
(iv) the exploitative behavior is indeed caused by the cross-hand encoder.

\subsection{Experimental setup}

AlphaExploitem is implemented in JAX~\citep{jax2018github} using Flax and Optax. 
The Kuhn and Leduc environments are provided by PGX~\citep{pgx}, a JAX-native library of game simulators enabling fully vectorized, GPU-accelerated trajectory generation. 
Training uses standard PPO~\citep{PPO} with AdamW, gradient clipping, KL-based early stopping.
Full architectural and hyperparameter details are given in Appendix~\ref{app:hyperparams}.

Each trained agent is evaluated in two modes: an exploiter mode with the full tokenized session history provided to the transformer, and a non-exploiter mode with the history masked so only the current-hand card and action encoders inform decisions.
Rewards are reported in big-blinds per hand (BBs/hand).
Two opponent pools are used for evaluation: the \textit{in-distribution} toy strategies seen during training and a held-out \textit{out-of-distribution} set of diverse hand-crafted policies. Information about individual toy policies can be found in Appendix~\ref{app:toy-strategies}.

\subsection{Training dynamics: exploitation ability}
\label{sec:results-evolution}

Our main claim is that the cross-hand encoder confers the main agent with the ability to extract additional EV (compared to the baseline) from exploitable opponents and that this advantage translates as well to the out-of-distribution (OOD) set.
We compare the main agent against two no-encoder baselines on both Kuhn and Leduc Hold'em: a pure league-self-play baseline (the AlphaHoldem-style setup with no toys and no snapshot buffer) and an ablation that isolates the contribution of the encoder by training the original AlphaHoldem architecture (no cross-hand encoder) using the new training paradigm.

\begin{figure}[H]
\centering
\includegraphics[width=\linewidth]{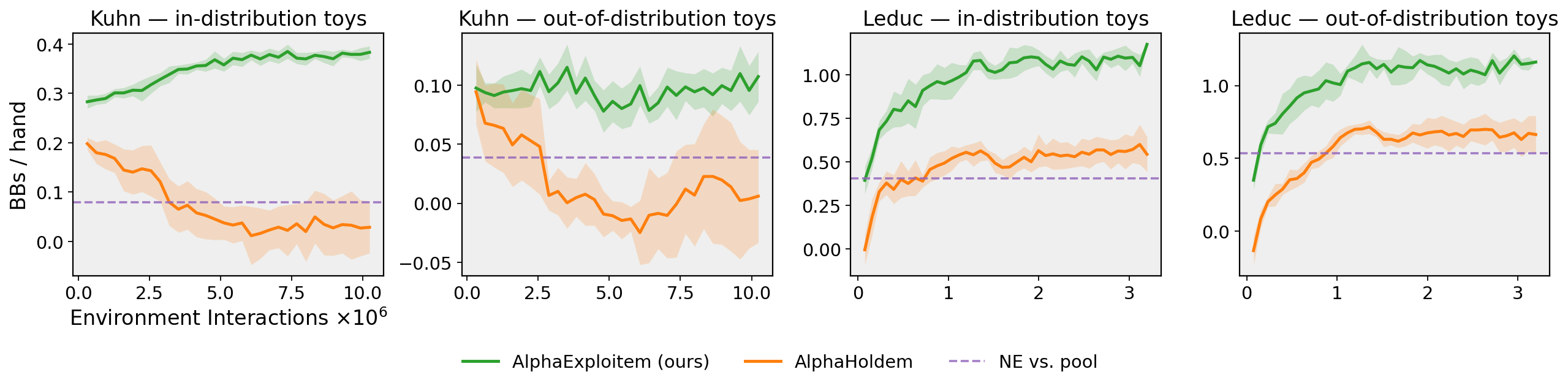}
\caption{Reward evolution against the in-distribution and out-of-distribution toy pools, on Kuhn Poker (left two panels) and Leduc Hold'em (right two panels). 95\% confidence intervals with 8 seeds. Horizontal axis: environment interactions (hands of play). The dashed line is the mean reward NE obtains against the corresponding pool (Appendix~\ref{app:nash-vs-toys}).}
\label{fig:evolution-4panel}
\end{figure}

Figure~\ref{fig:evolution-4panel} reports the seed-mean reward/hand against the in-distribution toy pool and a held-out out-of-distribution pool.
On Leduc the main agent dominates on both pools: it reaches an in-distribution (ID) reward of roughly $+1.0$ BBs/hand vs.\ $+0.5$ for AlphaHoldem.
The OOD ordering is identical.
The ID and OOD curves track one another closely throughout, indicating that the encoder learns class-level opponent features that transfer to held-out variants rather than overfitting per-toy fingerprints.
For Kuhn the in-distribution evolution shows the same qualitative pattern as Leduc: the main agent climbs to roughly $+0.38$ BBs/hand while AlphaHoldem stays close to zero ($\approx +0.03$).
The Kuhn OOD raw-reward curve, however, looks visually flat, which would naively suggest no improvement on held-out opponents.
This impression is misleading: per-toy reward divided by the toy's best-response (BR) ceiling rises monotonically with training (Figure~\ref{fig:brfrac-evolution-kuhn-ood} in Appendix~\ref{app:br-ceilings}).
The discrepancy is a result of unit choice in the averaging: a small absolute BB gain against a low-BR-ceiling toy registers strongly in BR-fraction terms but contributes negligibly to the chip-denominated mean, so improvements on the most exploitable Kuhn OOD toys are visible only after the per-toy normalisation.

\subsection{Baseline play vs.\ Nash}
\label{sec:results-nash}

\begin{figure}[H]
\centering
\includegraphics[width=\linewidth]{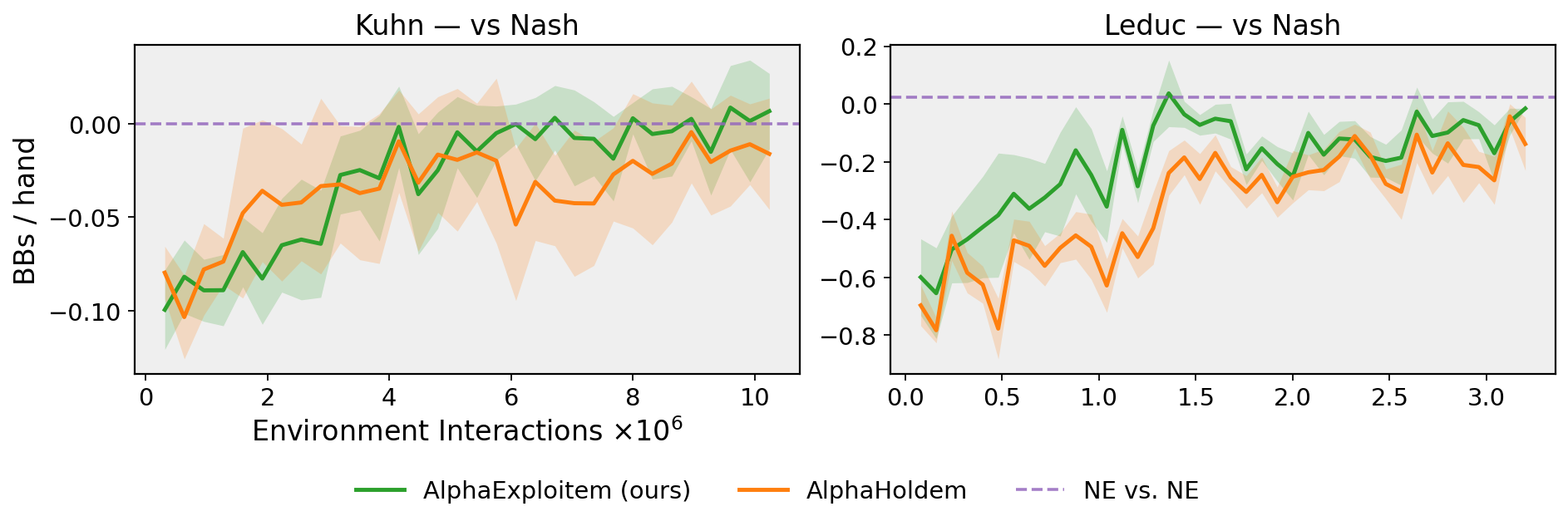}
\caption{Reward / hand vs.\ the Nash equilibrium policy on Kuhn (left) and Leduc Hold'em (right). 95\% confidence over 8 seeds. The dashed zero-line is the NE-vs-NE expectation, position-averaged.}
\label{fig:nash-2panel}
\end{figure}

We verify that the exploitation gains do not come at the cost of fundamental strength: a competent poker agent should hold its own against a Nash-equilibrium policy rather than be exploited by it. Figure~\ref{fig:nash-2panel} reports per-checkpoint mean BBs/hand against the (computed) Nash equilibrium.

On Leduc, both the baseline and AlphaExploitem improve steadily over training and converge close to the NE expectation. None of the two models reaches exact NE, but the gap shrinks monotonically with training and is meaningfully smaller than the gap to weak toy opponents.

\subsection{Per-opponent results}
\label{sec:results-per-toy}

\begin{figure}[H]
\centering
\includegraphics[width=0.92\linewidth]{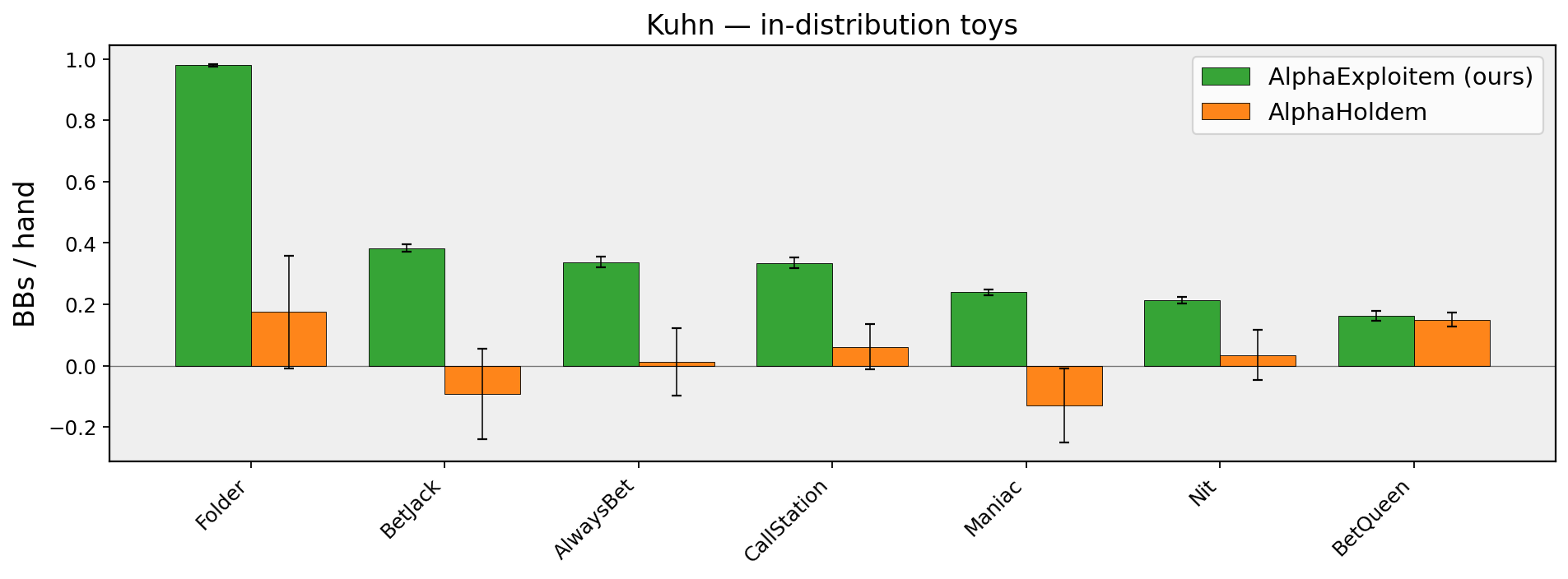}
\caption{Kuhn in-distribution final-checkpoint reward/hand against each toy opponent. Bars are seed-means over the last $5$ logged checkpoints with 95\% confidence error bars. N = 8 seeds per group. Toys are sorted left-to-right by descending average reward.}
\label{fig:per-toy-reward-kuhn-id}
\end{figure}

\begin{figure}[H]
\centering
\includegraphics[width=0.92\linewidth]{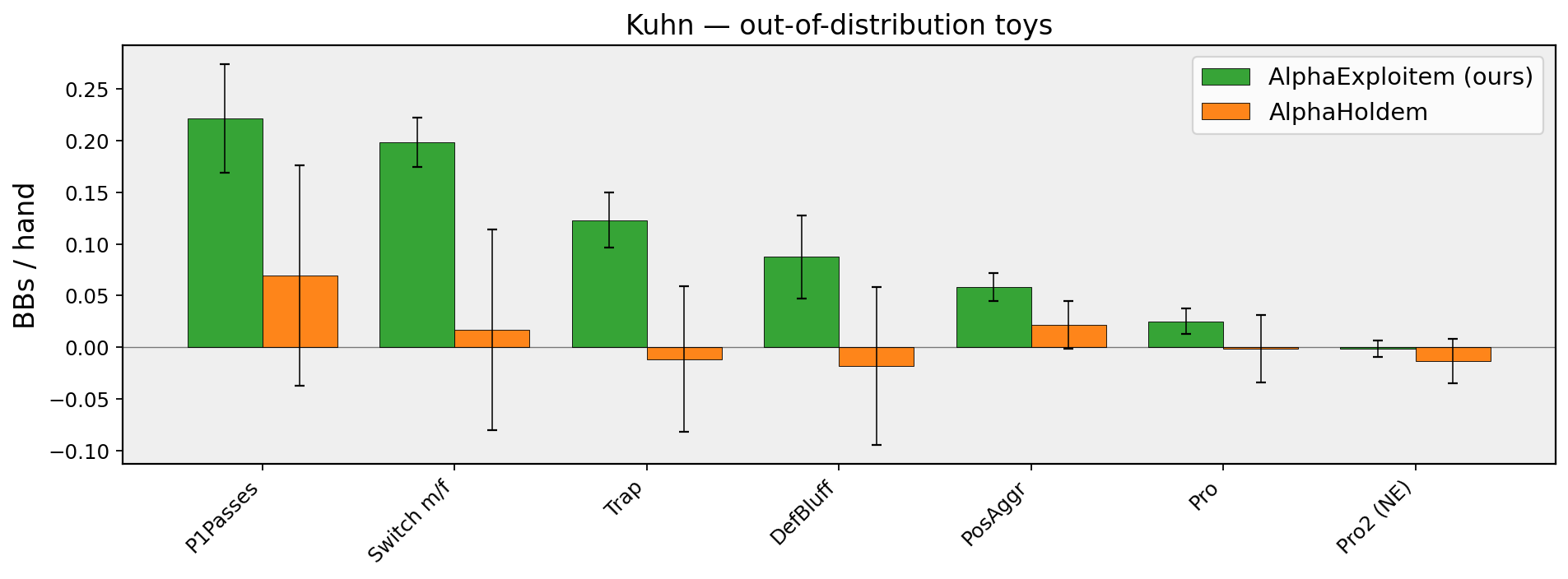}
\caption{Kuhn --- out-of-distribution toys. Same conventions as Figure~\ref{fig:per-toy-reward-kuhn-id}}
\label{fig:per-toy-reward-kuhn-ood}
\end{figure}

\begin{figure}[H]
\centering
\includegraphics[width=0.92\linewidth]{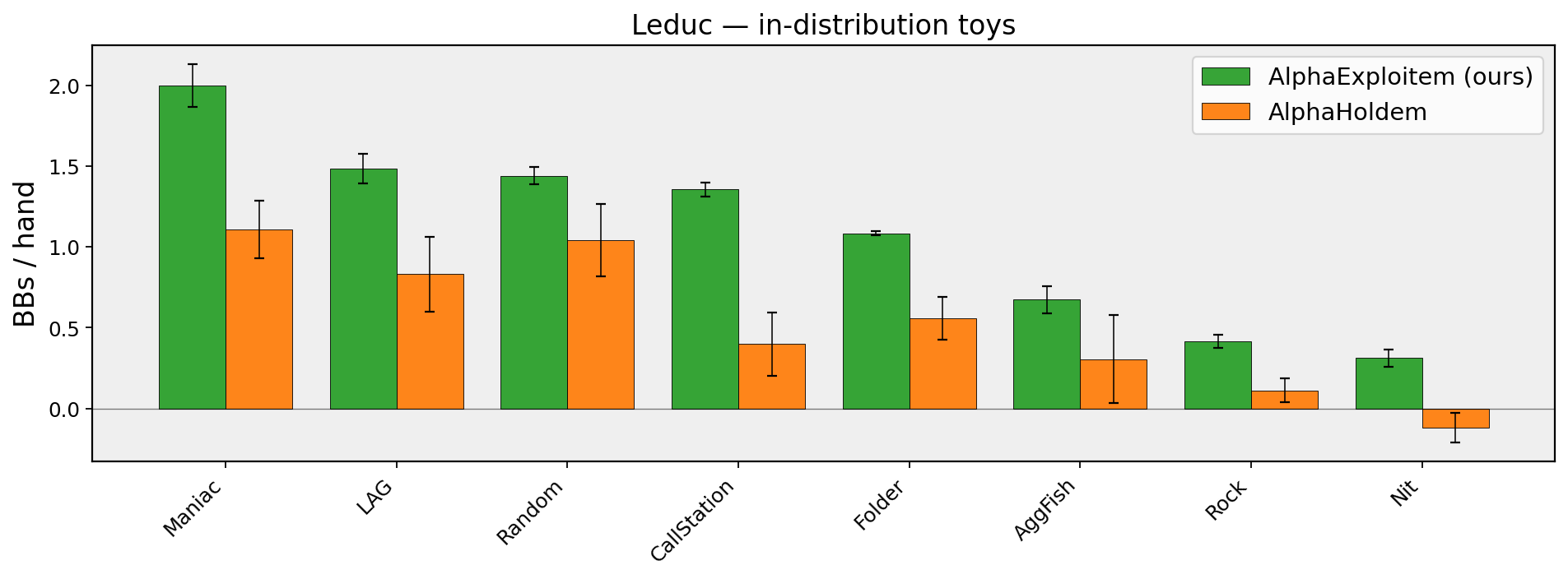}
\caption{Leduc --- in-distribution toys. Same conventions as Figure~\ref{fig:per-toy-reward-kuhn-id}}
\label{fig:per-toy-reward-leduc-id}
\end{figure}

\begin{figure}[H]
\centering
\includegraphics[width=0.92\linewidth]{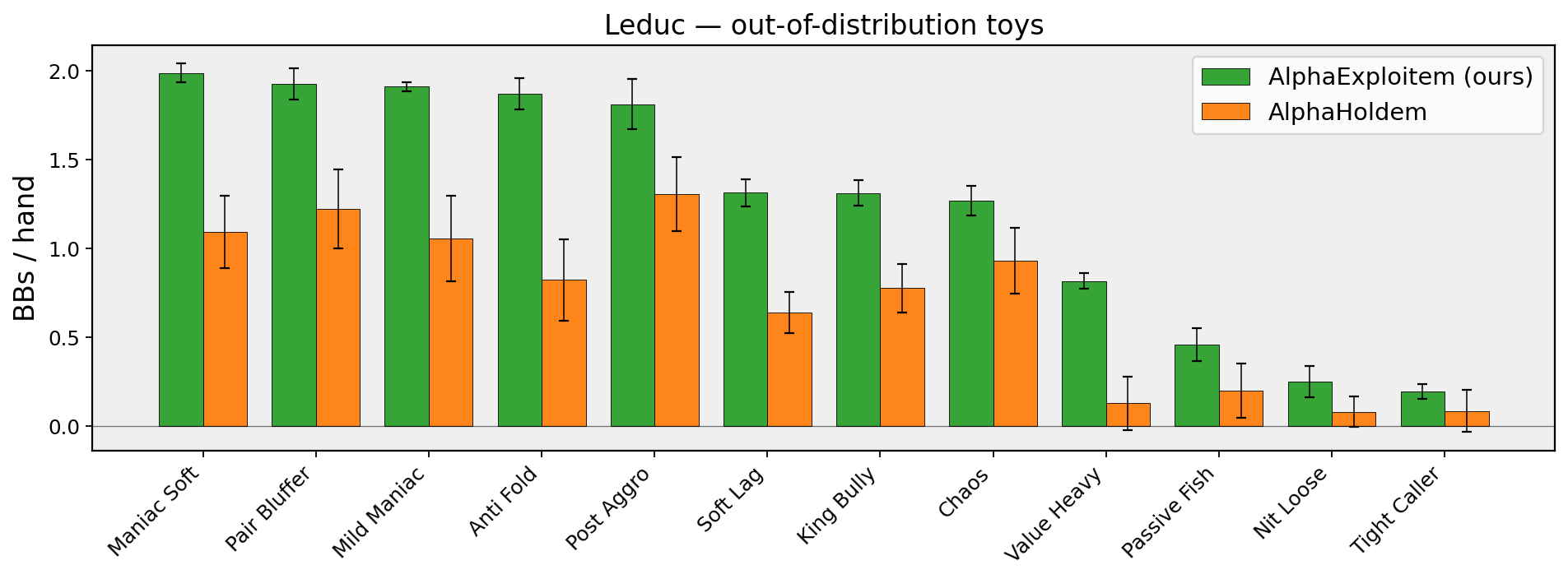}
\caption{Leduc --- out-of-distribution toys. Same conventions as Figure~\ref{fig:per-toy-reward-kuhn-id}}
\label{fig:per-toy-reward-leduc-ood}
\end{figure}

The pool-averaged curves of Section~\ref{sec:results-evolution} do not show whether a single dominant opponent is responsible for the gap. Figures~\ref{fig:per-toy-reward-kuhn-id}, \ref{fig:per-toy-reward-kuhn-ood}, \ref{fig:per-toy-reward-leduc-id}, \ref{fig:per-toy-reward-leduc-ood} decompose the rewards toy by toy, for both Kuhn and Leduc. The Appendix \ref{app:br-ceilings} variants (Figures~\ref{fig:per-toy-brfrac-kuhn-id}, \ref{fig:per-toy-brfrac-kuhn-ood}, \ref{fig:per-toy-brfrac-leduc-id}, \ref{fig:per-toy-brfrac-leduc-ood}) normalize the rewards by their Best Response (BR) ceiling in order to reveal exploit \emph{efficiency} rather than absolute reward.

The encoder advantage is broad rather than concentrated, and visible on both games. AlphaExploitem achieves better rewards on all toys. On Leduc the largest absolute gains line up with the most exploitable opponents --- maniac, loose aggressive (LAG), and the OOD bluffer toys. Cautious opponents like \texttt{nit} and \texttt{rock} leave little room to extract reward and the three groups bunch together near zero, again as expected from their small BR ceilings.

\subsection{Cross-hand context: masking ablation}
\label{sec:results-masking}

To isolate the contribution of context specifically, we evaluate AlphaExploitem against the same model but with the history completely masked during evaluation time so the policy must act on the current-hand card and action observations alone. Architecture, optimization, and league exposure are by construction identical between the two. The only difference is whether the transformer encoder sees the prior-hand tokens at decision time.

\begin{figure}[ht]
\centering
\includegraphics[width=\linewidth]{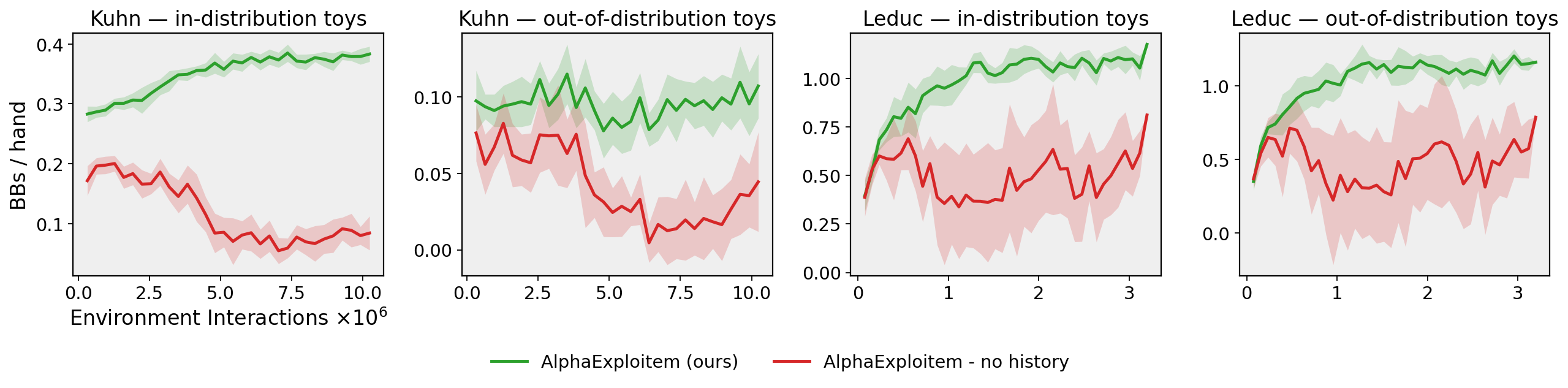}
\caption{Effect of masking the cross-hand history channel at evaluation time on the same trained AlphaExploitem policy. Four panels: Kuhn (left two) and Leduc (right two), each split into in-distribution and out-of-distribution toy pools. Per-group seed-mean is plotted with 95\% confidence over $8$ seeds.}
\label{fig:masking-ablation}
\end{figure}

Figure~\ref{fig:masking-ablation} reports the evolution of mean reward in both evaluation modes. The two lines start nearly together early in training, when the agent has not yet learned to use cross-hand context, and then diverge as the encoder learns to extract opponent-specific signal from the prior hands. The gap between the green and red curves is the contribution attributable specifically to in-context adaptation, with architecture, training, and parameters held fixed.

The size of the gap differs sharply between the two games. On \textbf{Leduc} the unmasked main agent earns $+1.10$ BBs/hand on the ID pool and $+1.16$ on the OOD pool (seed-means, $n{=}8$). Masking the same policy collapses these to $+0.54$ and $+0.52$ respectively. Roughly half of the agent's exploitation reward is attributable specifically to cross-hand context, and the OOD pool actually benefits slightly more from context than the ID pool ($\Delta_\text{OOD} = +0.64$ vs.\ $\Delta_\text{ID} = +0.55$ chips/hand). This is consistent with the encoder learning class-level opponent features rather than per-toy fingerprints. On \textbf{Kuhn} the gap is proportionally even larger: ID main reaches $+0.38$ vs.\ $+0.09$ masked, and OOD reaches $+0.10$ vs.\ $+0.03$ masked. The masked policy is barely better than zero on Kuhn OOD. This is intuitive given the game: a single Kuhn hand exposes very little hidden information about an opponent's style (only one card and a short betting line), so once context is removed the policy has almost nothing to specialise on.

A natural concern with masked-vs-unmasked attribution is that the masked variant is not constrained to play near-Nash. Nothing in training forces the same parameters to behave as a safe NE-approximator when context is removed, so the masked level is simply ``what this trained policy produces from card and current-hand action observations alone''. Two empirical observations bound this. First, on Leduc the masked aggregate sits noticeably above the (computed) NE level on both pools (Appendix~\ref{app:nash-vs-toys}), so the masked policy is on average more exploitative than safe NE play. Second, the masked aggregate is comparable to or below the no-encoder league-only baseline on the same pools, which suggests the masked policy is being overfit on the training set. We therefore read the masked-vs-unmasked gap as a \emph{lower bound} on the exploitation gain attributable to cross-hand context.

%% file: Sections/Conclusion.tex
\section{Conclusion}
\label{sec:conclusion}

We presented AlphaExploitem, a transformer-augmented poker agent that learns to exploit suboptimal opponents by processing multi-hand game histories through a lightweight transformer encoder. 
Built on AlphaHoldem's \citep{AlphaHoldem} architectural foundation and trained via PPO with K-best league self-play, the agent simultaneously learns exploitative adaptation and robust equilibrium-approximating play. 
We demonstrate the approach across Kuhn Poker and Leduc Hold'em, and present our findings regarding both exploitative decisions against flawed hand-crafted policies and robust play against Nash Equilibria.
Our results show that AlphaExploitem can learn to exploit a variety of opponent archetypes, achieving significant gains against hand-crafted policies while maintaining competitive performance against equilibrium strategies.

%% file: Sections/Appendices/Limitations.tex
\section{Limitations}
\label{app:lims}

This section discusses the limitations and the assumptions of our work, and what are the possible solutions that can improve our approach.

\textbf{Strong assumptions.} All the evaluation toy policies present stationary policies. The model is exposed to non-stationary policies during the league and the buffer play stages, but we do not integrate evaluation metrics for non-stationarity conditions. Additionally, we assume that AlphaExploitem plays matches against single opponents (heads-up play), with an unlimited supply of money but with a fixed amount of hands per game.

\textbf{Limitations.} While Leduc Hold'em presents the same characteristics as Texas Hold'em, it does not reach the same complexity. The toy crafting component of our approach may prove harder to reliably scale to such an environment. Another flaw of our model is its finite time-horizon. Alphaexploitem is not able to play infinitely without losing information about old hands. A possible solution for this would be to use a recurrent network with a hidden state that is updated by the latent dimensions of the cross-hand encoder.

\textbf{Computational Efficiency.} The proposed training algorithm inherits the parallelizability traits of the transformer~\citep{Transformers} architecture but also the data inefficiency of PPO~\citep{PPO}. For our experiments, we used a single NVIDIA A40 with a training time of 12 hours per Leduc seed and 4 hours per Kuhn seed.

%% file: Sections/Appendices/Player-Types.tex
\section{Player Types}
\label{app:players}

Real-world poker styles are conventionally placed on two axes: how often a player enters a hand (\textit{tight} vs.\ \textit{loose}) and how often they raise versus call (\textit{passive} vs.\ \textit{aggressive}).
The four resulting quadrants name the colloquial archetypes used throughout the poker community: \textit{tight-aggressive} (TAG, the canonical strong style), \textit{loose-aggressive} (LAG, profitable but high-variance), \textit{tight-passive} (the \textit{nit} or \textit{rock}, cautious to a fault), and \textit{loose-passive} (the \textit{calling station}, which calls down indiscriminately).
Two further endpoints sit at the extremes of the action distribution: the \textit{maniac} (raises and re-raises near-uniformly) and the \textit{folder} (folds whenever legal).
Except the TAG --- balanced play that an NE-approximating agent already handles --- each of these styles corresponds to a stationary mixed policy with predictable, exploitable structure: precisely the kind of suboptimal play our agent is trained to detect and exploit.
We instantiate the exploitable archetypes as fixed \textit{toy opponents} with explicit action-probability tables (Appendix~\ref{app:toy-strategies}), use them both during training and for in-distribution evaluation, and reserve unseen variants for the out-of-distribution evaluation pool.

%% file: Sections/Appendices/ToyStrategies.tex
\section{Toy strategies}
\label{app:toy-strategies}

This appendix specifies every toy opponent used in training and evaluation. All toys are stationary, history-(in)dependent mixed policies. A toy never learns or adapts. We give each toy's full action distribution as fixed probability tables.

\subsection{Kuhn poker toys}
\label{app:toy-kuhn}

Kuhn poker has three cards (J, Q, K) and two actions (BET, PASS). Each toy is a $3 \times 4$ matrix giving \textit{the probability of BET} given (hero card, betting history). Columns are the four possible histories at the hero's decision nodes: \textit{first to act}, \textit{after pass}, \textit{after bet}, \textit{after pass-bet}. ``After bet''/``After pass-bet'' rows are interpreted as the probability of betting (= calling). Subtracting from 1 gives the probability of passing (= folding).

\paragraph{In-distribution toys (training pool).} These seven toys are part of the training curriculum.

\begin{table}[H]
\centering
\caption{Kuhn ID toys: probability of BET. Cards: J=jack, Q=queen, K=king. Histories: \textit{(start)} = hero acts first. \textit{p} = opponent passed. \textit{b} = opponent bet. \textit{pb} = pass-bet line.}
\label{tab:kuhn-id-toys}
\small
\begin{tabular}{l c c c}
\toprule
\textbf{Toy} & $P(\text{BET}\mid J, \cdot)$ & $P(\text{BET}\mid Q, \cdot)$ & $P(\text{BET}\mid K, \cdot)$ \\
\midrule
\multicolumn{4}{l}{histories: $(\text{start},\ p,\ b,\ pb)$} \\
\midrule
\texttt{f}   (Folder)             & $(0.0,0.0,0.0,0.0)$ & $(0.0,0.0,0.0,0.0)$ & $(0.0,0.0,0.0,0.0)$ \\
\texttt{abj} (AlwaysBetJacks)     & $(1.0,1.0,0.0,0.0)$ & $(0.0,0.0,0.5,0.0)$ & $(0.0,0.0,1.0,1.0)$ \\
\texttt{ab}  (AlwaysBet)          & $(1.0,1.0,1.0,0.0)$ & $(1.0,1.0,1.0,0.0)$ & $(1.0,1.0,1.0,0.0)$ \\
\texttt{cs}  (Calling Station)    & $(0.0,0.0,0.5,1.0)$ & $(0.0,0.0,1.0,1.0)$ & $(1.0,1.0,1.0,1.0)$ \\
\texttt{m}   (Maniac)             & $(1.0,1.0,0.0,0.0)$ & $(0.0,0.0,1.0,1.0)$ & $(1.0,1.0,1.0,1.0)$ \\
\texttt{n}   (Nit)                & $(0.0,0.0,0.0,0.0)$ & $(0.0,0.0,0.0,0.0)$ & $(1.0,1.0,1.0,1.0)$ \\
\texttt{abq} (AlwaysBetQueens)    & $(0.0,0.0,0.0,0.0)$ & $(1.0,1.0,1.0,1.0)$ & $(0.0,0.0,0.6,0.6)$ \\
\bottomrule
\end{tabular}
\end{table}

\paragraph{Out-of-distribution toys (evaluation only).} These seven toys are never sampled during training.

\begin{table}[H]
\centering
\caption{Kuhn OOD toys: probability of BET. Layout matches Table~\ref{tab:kuhn-id-toys}. \texttt{ood\_switch\_mf} is non-stationary: it alternates between Maniac and Folder every $50$ hands within a session, so the entries below summarise its two component policies.}
\label{tab:kuhn-ood-toys}
\small
\begin{tabular}{l c c c}
\toprule
\textbf{Toy} & $P(\text{BET}\mid J,\cdot)$ & $P(\text{BET}\mid Q,\cdot)$ & $P(\text{BET}\mid K,\cdot)$ \\
\midrule
\texttt{ood\_p1p}       (P1Passes)        & $(0.1,0.33,0.0,0.0)$ & $(0.1,0.0,0.33,0.0)$ & $(0.1,1.0,1.0,0.0)$ \\
\texttt{ood\_switch\_mf}\;(Maniac half)   & $(1.0,1.0,0.0,0.0)$  & $(0.0,0.0,1.0,1.0)$  & $(1.0,1.0,1.0,1.0)$ \\
\texttt{ood\_switch\_mf}\;(Folder half)   & $(0.0,0.0,0.0,0.0)$  & $(0.0,0.0,0.0,0.0)$  & $(0.0,0.0,0.0,0.0)$ \\
\texttt{ood\_trap}      (Trap)            & $(0.5,0.5,0.0,0.0)$  & $(0.7,0.5,0.4,0.3)$  & $(0.0,1.0,1.0,1.0)$ \\
\texttt{ood\_def\_bluff}(DefBluff)        & $(0.0,0.6,0.0,0.0)$  & $(0.0,0.4,0.0,0.0)$  & $(0.7,1.0,1.0,1.0)$ \\
\texttt{ood\_pos\_aggr} (PosAggr)         & $(0.4,0.0,0.0,0.0)$  & $(0.5,0.1,0.5,0.4)$  & $(0.9,0.5,1.0,1.0)$ \\
\texttt{ood\_p}         (Pro)             & $(0.1,0.33,0.0,0.0)$ & $(0.33,0.0,0.33,0.33)$ & $(1.0,1.0,1.0,1.0)$ \\
\texttt{ood\_p2}        (Pro2 / NE)       & $(0.1,0.33,0.0,0.0)$ & $(0.0,0.0,0.33,0.43)$ & $(0.3,1.0,1.0,1.0)$ \\
\bottomrule
\end{tabular}
\end{table}

\subsection{Leduc Hold'em toys}
\label{app:toy-leduc}

Leduc has six cards (J, Q, K $\times$ 2 suits but suits are irrelevant for resolution), two betting rounds, and three actions (CALL/CHECK, BET/RAISE, FOLD). Toys condition on \emph{(hero rank, community rank, round)} but not on the hero's preceding actions. Some additionally condition on whether a bet is pending (i.e.\ on the legal-action mask, which gates FOLD). All probabilities below are renormalized over legal actions at runtime.

\paragraph{In-distribution toys (training and evaluation pool).} Eight toys, sampled uniformly within each training batch.

\begin{table}[H]
\centering
\caption{Leduc ID toys. Action probabilities are ordered as $(\text{call/check},\,\text{bet/raise},\,\text{fold})$. Where the policy varies with state we list each branch separately. Otherwise the toy plays a single fixed mixture.}
\label{tab:leduc-id-toys}
\small
\begin{tabular}{l p{0.62\linewidth}}
\toprule
\texttt{maniac} & $(0.05, 0.95, 0.0)$ everywhere — aggressive toy. \\
\midrule
\texttt{lag} &
Pre-flop: $(0.25, 0.7, 0.05)$. Post-flop pair: $(0.05, 0.95, 0.0)$. Post-flop no-pair: $(0.3, 0.55, 0.15)$. \\
\midrule
\texttt{random} & Uniform random over the legal actions at every decision. \\
\midrule
\texttt{calling\_station} & $(1.0, 0.0, 0.0)$ everywhere — pure call. \\
\midrule
\texttt{folder} & $(0.05, 0.0, 0.95)$ everywhere — cautious toy. \\
\midrule
\texttt{aggfish} &
Pre-flop: $(0.1, 0.9, 0.0)$. Post-flop pair: $(0.05, 0.95, 0.0)$. Post-flop no-pair: $(0.1, 0.05, 0.85)$. \\
\midrule
\texttt{rock} &
Pre-flop: J $(0.3, 0.0, 0.7)$, Q/K $(0.95, 0.0, 0.05)$. Post-flop pair $(0.8, 0.15, 0.05)$, K-no-pair $(0.6, 0.0, 0.4)$, otherwise $(0.1, 0.0, 0.9)$. \\
\midrule
\texttt{nit} &
Pre-flop: J $(0.0,0.0,1.0)$, Q $(0.8,0.0,0.2)$, K $(0.3,0.7,0.0)$. Post-flop pair $(0.3,0.7,0.0)$, K-no-pair $(0.7,0.1,0.2)$, otherwise $(0.0,0.0,1.0)$. \\
\bottomrule
\end{tabular}
\end{table}

\paragraph{Out-of-distribution toys (evaluation only).} Twelve toys never sampled during training. They are novel combinations or perturbations of the styles seen in training.

\begin{table}[H]
\centering
\caption{Leduc OOD toys. Layout matches Table~\ref{tab:leduc-id-toys}. Action probabilities $(\text{call/check},\,\text{bet/raise},\,\text{fold})$.}
\label{tab:leduc-ood-toys}
\small
\begin{tabular}{l p{0.62\linewidth}}
\toprule
\texttt{ood\_maniac\_soft}     & $(0.2, 0.75, 0.05)$ — softer maniac. \\
\midrule
\texttt{ood\_pair\_bluffer}    & Pre-flop $(0.4,0.5,0.1)$. Post-flop pair $(0.85,0.1,0.05)$ (trap), no-pair $(0.1,0.8,0.1)$ (bluff). Inverted value-betting. \\
\midrule
\texttt{ood\_mild\_maniac}     & $(0.3, 0.6, 0.1)$ — between LAG and Maniac. \\
\midrule
\texttt{ood\_anti\_fold}       & $(0.5,0.5,0.0)$ — never folds. \\
\midrule
\texttt{ood\_post\_aggro}      & Pre-flop $(0.9,0.05,0.05)$. Post-flop $(0.15,0.7,0.15)$ regardless of card. \\
\midrule
\texttt{ood\_soft\_lag}        & Pre-flop $(0.45,0.45,0.1)$. Post-flop pair $(0.15,0.85,0.0)$, no-pair $(0.45,0.35,0.2)$. \\
\midrule
\texttt{ood\_king\_bully}      & K $(0.05,0.9,0.05)$, J/Q $(0.8,0.05,0.15)$. \\
\midrule
\texttt{ood\_chaos}            & Per-decision: with prob $1/3$ commit to all-call, $1/3$ all-raise, $1/3$ all-fold. \\
\midrule
\texttt{ood\_value\_heavy}     & Pre-flop $(0.7,0.2,0.1)$. Post-flop pair $(0.1,0.9,0.0)$, no-pair $(0.8,0.0,0.2)$. \\
\midrule
\texttt{ood\_passive\_fish}    & Pre-flop $(0.85,0.1,0.05)$. Post-flop pair $(0.4,0.6,0.0)$, no-pair $(0.15,0.0,0.85)$. \\
\midrule
\texttt{ood\_nit\_loose}       & Pre-flop: J $(0.2,0.0,0.8)$, Q/K $(0.5,0.4,0.1)$. Post-flop pair $(0.2,0.8,0.0)$, Q/K no-pair $(0.5,0.1,0.4)$, otherwise $(0.0,0.0,1.0)$. \\
\midrule
\texttt{ood\_tight\_caller}    & Pre-flop: J $(0.0,0.0,1.0)$, Q/K $(0.95,0.0,0.05)$. Post-flop pair $(0.9,0.05,0.05)$, J $(0.0,0.0,1.0)$, Q/K no-pair $(0.8,0.0,0.2)$. \\
\bottomrule
\end{tabular}
\end{table}

%% file: Sections/Appendices/NashEquilibria.tex
\section{Nash equilibria}
\label{app:nash}

This appendix records the Nash-equilibrium strategies used as the reference policy in the vs-Nash evaluation of Section~\ref{sec:results-nash}. For Kuhn the equilibrium is a one-parameter family with a closed form. For Leduc Hold'em it is a sampled finite table over the game's information sets.

\subsection{Kuhn Nash equilibrium}
\label{app:nash-kuhn}

Kuhn poker admits a one-parameter family of Nash-equilibrium strategies indexed by $\alpha \in [0, 1/3]$~\citep{kuhn2016simplified}. Both players' strategies are listed below. Cards are J=jack, Q=queen, K=king. Actions are BET and PASS. The betting trees considered are \textit{(start)} (the first to act), \textit{after pass}, \textit{after bet}, and \textit{after pass-bet}.

\begin{table}[H]
\centering
\caption{Player 1 (first to act).}
\label{tab:nash-kuhn-p1}
\begin{tabular}{l c c c}
\toprule
history & $J$ & $Q$ & $K$ \\
\midrule
\textit{(start)} & $\alpha$ & $0$            & $3\alpha$ \\
\textit{p}       & $0$      & $0$            & $1$ \\
\textit{b}       & $0$      & $1/3 + \alpha$ & $1$ \\
\textit{pb}      & $0$      & $1/3 + \alpha$ & $1$ \\
\bottomrule
\end{tabular}
\end{table}

\begin{table}[H]
\centering
\caption{Player 2 (second to act).}
\label{tab:nash-kuhn-p2}
\begin{tabular}{l c c c}
\toprule
history & $J$ & $Q$ & $K$ \\
\midrule
\textit{p} & $1/3$ & $0$   & $1$ \\
\textit{b} & $0$   & $1/3$ & $1$ \\
\bottomrule
\end{tabular}
\end{table}

\paragraph{Game value.} Under any Nash strategy in this family the game value is $V_1 = -1/18$ chips per hand for Player 1 (in position. First to act preflop) and $+1/18$ for Player 2. The vs-Nash evaluation in Section~\ref{sec:results-nash} reports rewards averaged across positions (since the dealer is randomized per hand by the simulator), so the relevant zero-line is $0$ chips/hand and not $-1/18$.

For our evaluations we instantiate $\alpha = 1/3$ as a canonical choice.

\subsection{Leduc Hold'em Nash equilibrium}
\label{app:nash-leduc}

Leduc Hold'em has a much larger but still finite information-set space. Our reference Nash strategy is a tabulated equilibrium with one row per information set, computed via counterfactual regret minimisation to convergence. We expose two slices below.

\paragraph{Round-1 information sets.} Round 1 (pre-community card) has 18 reachable information sets keyed by the player's own card and the action history so far. Action letters: $x = $ check, $b = $ bet, $r = $ raise, $c = $ call, $f = $ fold. Each row sums to 1.

\begin{table}[H]
\centering
\caption{Leduc Nash strategy, round-1 slice. Rows are info sets (\textit{card history}). Columns are action probabilities. Entries below $10^{-4}$ are reported as $0.000$. Game value: $V_1 \approx -0.0856$ chips per hand for Player 1 (first to act preflop), $+0.0856$ for Player 2.}
\label{tab:nash-leduc-r1}
\small
\begin{tabular}{l c c c c c}
\toprule
info set & $x$ (check) & $b$ (bet) & $r$ (raise) & $c$ (call) & $f$ (fold) \\
\midrule
J            & $0.926$ & $0.074$ & -     & -     & -     \\
J~b          & -     & -     & $0.055$ & $0.126$ & $0.819$ \\
J~br         & -     & -     & -     & $1.000$ & $0.000$ \\
J~x          & $0.704$ & $0.296$ & -     & -     & -     \\
J~xb         & -     & -     & $0.018$ & $0.035$ & $0.947$ \\
J~xbr        & -     & -     & -     & $1.000$ & $0.000$ \\
\midrule
Q            & $0.256$ & $0.744$ & -     & -     & -     \\
Q~b          & -     & -     & $0.369$ & $0.631$ & $0.000$ \\
Q~br         & -     & -     & -     & $1.000$ & $0.000$ \\
Q~x          & $0.151$ & $0.849$ & -     & -     & -     \\
Q~xb         & -     & -     & $0.160$ & $0.840$ & $0.000$ \\
Q~xbr        & -     & -     & -     & $1.000$ & $0.000$ \\
\midrule
K            & $0.246$ & $0.754$ & -     & -     & -     \\
K~b          & -     & -     & $0.580$ & $0.420$ & $0.000$ \\
K~br         & -     & -     & -     & $1.000$ & $0.000$ \\
K~x          & $0.000$ & $1.000$ & -     & -     & -     \\
K~xb         & -     & -     & $0.688$ & $0.312$ & $0.000$ \\
K~xbr        & -     & -     & -     & $1.000$ & $0.000$ \\
\bottomrule
\end{tabular}
\end{table}

\paragraph{Round-2 information sets.} After the community card is dealt, each round-1 reachable history can be extended by a community card $\in \{J, Q, K\}$ and a fresh round-2 betting line. With three card configurations $\times$ six round-1 histories $\times$ six round-2 betting trees, the round-2 table has roughly 100 rows. We omit the full listing here for space. The qualitative pattern is intuitive: with a pair the policy bets aggressively (e.g.\ $P(b \mid \texttt{JJ xxd}) \approx 1.0$), with no pair and against a bet it folds or bluff-raises depending on hand strength relative to the community card.

\subsection{NE evaluation against the toy pools}
\label{app:nash-vs-toys}

The NE policies of Sections~\ref{app:nash-kuhn}--\ref{app:nash-leduc} themselves obtain a non-zero reward against the suboptimal toys, since NE play is safe-but-not-exploitative. This provides the reference value used in the dashed lines in Figure~\ref{fig:evolution-4panel}. Tables~\ref{tab:ne-vs-toys-kuhn} and~\ref{tab:ne-vs-toys-leduc} report the per-toy and aggregate NE reward, split by which seat NE occupies. Kuhn values are exact (closed-form EV over the $12$-info-set tree). Leduc values are Monte-Carlo over $20{,}000$ hands per matchup, split evenly across the two seat assignments, against the NE policy.

\begin{table}[H]
\centering
\caption{NE vs.\ toy reward on Kuhn (chips/hand). Columns are NE's reward when it acts as P0 (first to act), as P1 (second to act), and the seat-averaged mean used as the dashed purple reference in Figure~\ref{fig:evolution-4panel}. Toys within each pool are sorted by descending mean.}
\label{tab:ne-vs-toys-kuhn}
\small
\begin{tabular}{l c c c}
\toprule
\textbf{Opponent} & NE-is-P0 & NE-is-P1 & Mean \\
\midrule
\multicolumn{4}{l}{\textit{In-distribution toys}} \\
\midrule
\texttt{abq}  & $+0.1533$ & $+0.1778$ & $+0.1656$ \\
\texttt{f}    & $+0.0667$ & $+0.2222$ & $+0.1444$ \\
\texttt{ab}   & $+0.1111$ & $+0.1111$ & $+0.1111$ \\
\texttt{cs}   & $-0.0306$ & $+0.2222$ & $+0.0958$ \\
\texttt{abj}  & $+0.0167$ & $+0.0556$ & $+0.0361$ \\
\texttt{m}    & $-0.0556$ & $+0.0556$ & $\phantom{+}0.0000$ \\
\texttt{n}    & $-0.0556$ & $+0.0556$ & $\phantom{+}0.0000$ \\
\midrule
\textbf{ID aggregate}    & - & - & $\mathbf{+0.0790}$ \\
\midrule
\multicolumn{4}{l}{\textit{Out-of-distribution toys}} \\
\midrule
\texttt{ood\_p1p}        & $-0.0556$ & $+0.2111$ & $+0.0778$ \\
\texttt{ood\_switch\_mf} & $+0.0056$ & $+0.1389$ & $+0.0722$ \\
\texttt{ood\_trap}       & $+0.0028$ & $+0.0944$ & $+0.0486$ \\
\texttt{ood\_pos\_aggr}  & $-0.0078$ & $+0.0833$ & $+0.0378$ \\
\texttt{ood\_def\_bluff} & $-0.0089$ & $+0.0556$ & $+0.0233$ \\
\texttt{ood\_p}          & $-0.0556$ & $+0.0741$ & $+0.0093$ \\
\texttt{ood\_p2}         & $-0.0556$ & $+0.0556$ & $\phantom{+}0.0000$ \\
\midrule
\textbf{OOD aggregate}   & - & - & $\mathbf{+0.0384}$ \\
\bottomrule
\end{tabular}
\end{table}

\begin{table}[H]
\centering
\caption{NE vs.\ toy reward on Leduc Hold'em (chips/hand). Columns are NE's reward when it acts as P1 (first to act), as P2 (second to act), and the seat-averaged mean. Monte-Carlo over $20{,}000$ hands per matchup.}
\label{tab:ne-vs-toys-leduc}
\small
\begin{tabular}{l c c c}
\toprule
\textbf{Opponent} & NE-is-P1 & NE-is-P2 & Mean \\
\midrule
\multicolumn{4}{l}{\textit{In-distribution toys}} \\
\midrule
\texttt{random}            & $+0.5270$ & $+0.7896$ & $+0.6583$ \\
\texttt{calling\_station}  & $+0.5254$ & $+0.7535$ & $+0.6394$ \\
\texttt{folder}            & $+0.5004$ & $+0.6583$ & $+0.5794$ \\
\texttt{lag}               & $+0.4665$ & $+0.6083$ & $+0.5374$ \\
\texttt{maniac}            & $+0.1242$ & $+0.6082$ & $+0.3662$ \\
\texttt{aggfish}           & $+0.1847$ & $+0.2584$ & $+0.2215$ \\
\texttt{rock}              & $+0.1127$ & $+0.3251$ & $+0.2189$ \\
\texttt{nit}               & $-0.0463$ & $+0.1096$ & $+0.0316$ \\
\midrule
\textbf{ID aggregate}      & - & - & $\mathbf{+0.4066}$ \\
\midrule
\multicolumn{4}{l}{\textit{Out-of-distribution toys}} \\
\midrule
\texttt{ood\_chaos}             & $+0.6344$ & $+0.8816$ & $+0.7580$ \\
\texttt{ood\_anti\_fold}        & $+0.5585$ & $+0.8463$ & $+0.7024$ \\
\texttt{ood\_post\_aggro}       & $+0.6531$ & $+0.6355$ & $+0.6443$ \\
\texttt{ood\_mild\_maniac}      & $+0.6070$ & $+0.6279$ & $+0.6175$ \\
\texttt{ood\_soft\_lag}         & $+0.4753$ & $+0.7461$ & $+0.6107$ \\
\texttt{ood\_king\_bully}       & $+0.4257$ & $+0.7619$ & $+0.5938$ \\
\texttt{ood\_pair\_bluffer}     & $+0.5842$ & $+0.6020$ & $+0.5931$ \\
\texttt{ood\_maniac\_soft}      & $+0.4312$ & $+0.6871$ & $+0.5592$ \\
\texttt{ood\_value\_heavy}      & $+0.2817$ & $+0.8176$ & $+0.5496$ \\
\texttt{ood\_passive\_fish}     & $+0.1790$ & $+0.4847$ & $+0.3318$ \\
\texttt{ood\_tight\_caller}     & $+0.1293$ & $+0.4064$ & $+0.2678$ \\
\texttt{ood\_nit\_loose}        & $+0.0692$ & $+0.3123$ & $+0.1908$ \\
\midrule
\textbf{OOD aggregate} & - & - & $\mathbf{+0.5349}$ \\
\bottomrule
\end{tabular}
\end{table}

\paragraph{Reading the tables.} The per-seat split exposes a substantial asymmetry: Leduc NE extracts noticeably more reward when it acts second (the Maniac column is the extreme: $+0.124$ as P1, $+0.608$ as P2, since acting after a maniac's bet is much more profitable than acting before). The aggregate values --- $+0.0790$ (Kuhn ID), $+0.0384$ (Kuhn OOD), $+0.4066$ (Leduc ID), $+0.5349$ (Leduc OOD) --- are the dashed references in Figure~\ref{fig:evolution-4panel} and bound the exploitation regime: a trained agent below the line is being out-exploited by a non-adaptive NE policy on the same pool.

%% file: Sections/Appendices/BestResponseCeilings.tex
\section{Best-response ceilings}
\label{app:br-ceilings}

The \textbf{best-response (BR) ceiling} of a (deterministic or mixed) opponent policy $\pi^o$ is the value of the BR policy against $\pi^o$:
\[
R_\text{BR}(\pi^o) \;=\; \max_{\pi^h}\, \mathbb{E}\bigl[\text{hero reward per hand} \,\big|\, \pi^h \text{ vs.}\ \pi^o\bigr].
\]
This is the ``oracle'' upper bound on what \emph{any} hero policy with full knowledge of $\pi^o$ can extract per hand, and it lets us report a unit-free \textbf{BR-fraction} $R / R_\text{BR}(\pi^o) \in [0, 1]$ alongside the raw reward, which is comparable across opponents of very different exploitability.

\paragraph{Leduc per-opponent values.} Table~\ref{tab:br-ceilings} reports $R_\text{BR}(\pi^o)$ in chips per hand for every Leduc toy used in the paper. Toys are sorted within each pool by descending ceiling --- i.e.\ how much exploit potential a perfect adversary could extract. This ordering is the ``$T_n$'' index used in the per-opponent figures of Section~\ref{sec:results-evolution}.

\begin{table}[H]
\centering
\caption{Best-response ceilings $R_\text{BR}(\pi^o)$ for the Leduc toy pools, in BBs/hand. Higher = more exploitable. ID toys are present in the training curriculum. OOD toys are evaluation-only.}
\label{tab:br-ceilings}
\small
\begin{tabular}{l c | l c}
\toprule
\multicolumn{2}{c|}{\textbf{In-distribution (ID)}} & \multicolumn{2}{c}{\textbf{Out-of-distribution (OOD)}} \\
toy & $R_\text{BR}$ & toy & $R_\text{BR}$ \\
\midrule
\texttt{random}           & $2.375$ & \texttt{ood\_pair\_bluffer}      & $3.208$ \\
\texttt{maniac}           & $2.366$ & \texttt{ood\_post\_aggro}        & $3.139$ \\
\texttt{aggfish}          & $1.946$ & \texttt{ood\_maniac\_soft}       & $2.423$ \\
\texttt{lag}              & $1.829$ & \texttt{ood\_mild\_maniac}       & $2.285$ \\
\texttt{calling\_station} & $1.467$ & \texttt{ood\_king\_bully}        & $2.213$ \\
\texttt{folder}           & $1.093$ & \texttt{ood\_anti\_fold}         & $2.149$ \\
\texttt{nit}              & $0.604$ & \texttt{ood\_chaos}              & $1.844$ \\
\texttt{rock}             & $0.520$ & \texttt{ood\_soft\_lag}          & $1.561$ \\
                          &         & \texttt{ood\_value\_heavy}       & $1.368$ \\
                          &         & \texttt{ood\_passive\_fish}      & $1.250$ \\
                          &         & \texttt{ood\_tight\_caller}      & $0.842$ \\
                          &         & \texttt{ood\_nit\_loose}         & $0.754$ \\
\bottomrule
\end{tabular}
\end{table}

\paragraph{Kuhn per-opponent values.} Table~\ref{tab:br-ceilings-kuhn} reports the analogous BR ceilings for the Kuhn toy pools. Values are exact (closed-form expectimax over the $12$-info-set hero tree). The folder \texttt{f} sits at the absolute exploitation ceiling of $1$~BB/hand because a perfect BR can simply bet every hand and win the starting pot uncontested the inverted-value \texttt{abj} and the indiscriminate \texttt{cs} carry the next-largest ceilings. The two near-Nash baselines (\texttt{ood\_p}, \texttt{ood\_p2}) sit near $0$ as expected.

\begin{table}[H]
\centering
\caption{Best-response ceilings $R_\text{BR}(\pi^o)$ for the Kuhn toy pools, in chips per hand, computed exactly by closed-form expectimax. Higher = more exploitable. ID toys are present in the training curriculum. OOD toys are evaluation-only.}
\label{tab:br-ceilings-kuhn}
\small
\begin{tabular}{l c | l c}
\toprule
\multicolumn{2}{c|}{\textbf{In-distribution (ID)}} & \multicolumn{2}{c}{\textbf{Out-of-distribution (OOD)}} \\
toy & $R_\text{BR}$ & toy & $R_\text{BR}$ \\
\midrule
\texttt{f}    & $1.000$ & \texttt{ood\_p1p}        & $0.439$ \\
\texttt{abj}  & $0.417$ & \texttt{ood\_switch\_mf} & $0.333$ \\
\texttt{cs}   & $0.375$ & \texttt{ood\_def\_bluff} & $0.275$ \\
\texttt{ab}   & $0.333$ & \texttt{ood\_trap}       & $0.191$ \\
\texttt{m}    & $0.250$ & \texttt{ood\_pos\_aggr}  & $0.117$ \\
\texttt{n}    & $0.250$ & \texttt{ood\_p}          & $0.066$ \\
\texttt{abq}  & $0.217$ & \texttt{ood\_p2}         & $0.001$ \\
\bottomrule
\end{tabular}
\end{table}

\begin{figure}[H]
\centering
\includegraphics[width=0.92\linewidth]{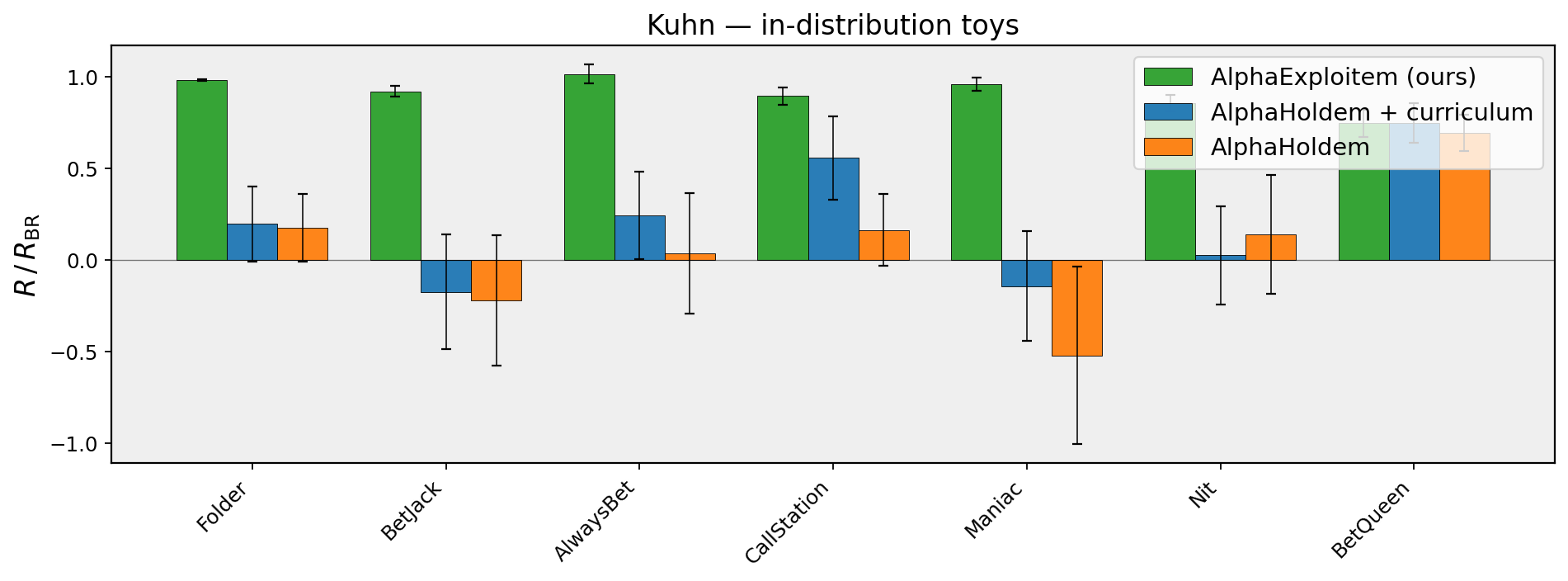}
\caption{Kuhn in-distribution final-checkpoint $R/R_\text{BR}$ against each toy opponent. Bars are seed-means over the last $5$ logged checkpoints with 95\% confidence error bars. N = 8 seeds per group. Toys are sorted left-to-right by descending average reward.}
\label{fig:per-toy-brfrac-kuhn-id}
\end{figure}

\begin{figure}[H]
\centering
\includegraphics[width=0.92\linewidth]{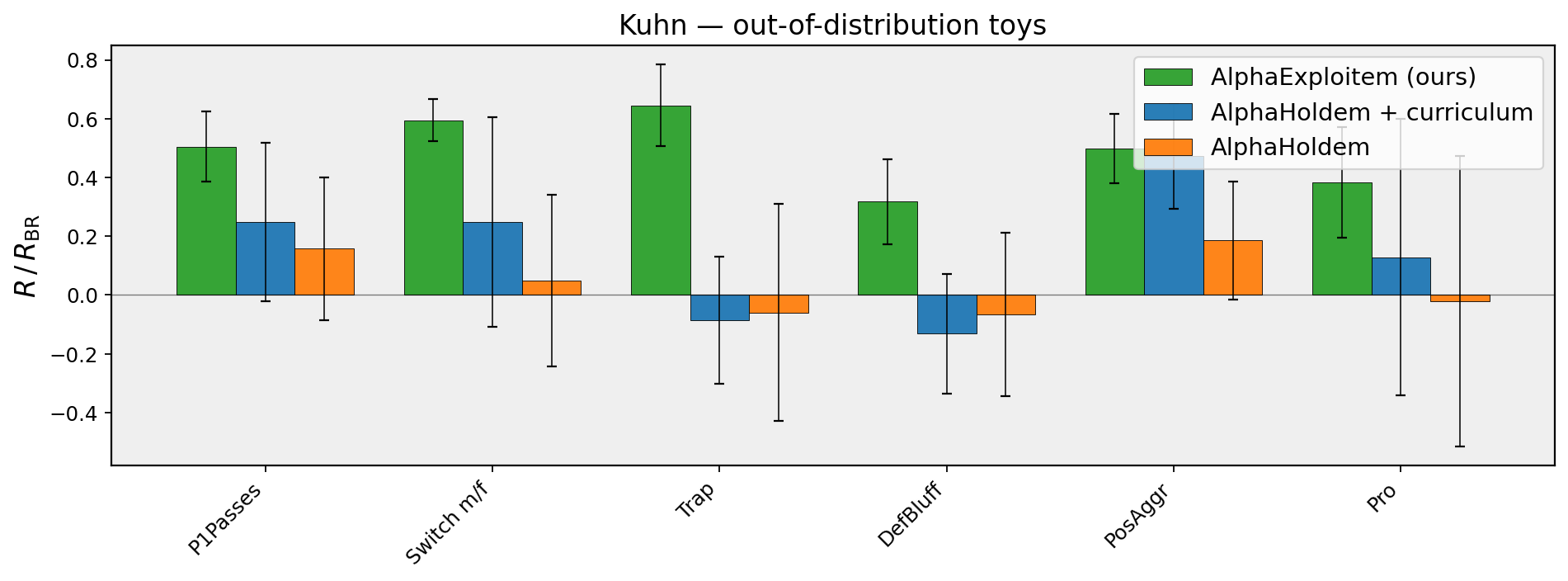}
\caption{Kuhn --- out-of-distribution toys. Same conventions as Figure~\ref{fig:per-toy-brfrac-kuhn-id}. \texttt{ood\_p2} (Pro2 / Nash) is omitted: its BR ceiling is $\approx 0.001$~BB/hand by construction, so the BR-fraction is dominated by noise rather than agent quality.}
\label{fig:per-toy-brfrac-kuhn-ood}
\end{figure}

\begin{figure}[H]
\centering
\includegraphics[width=0.92\linewidth]{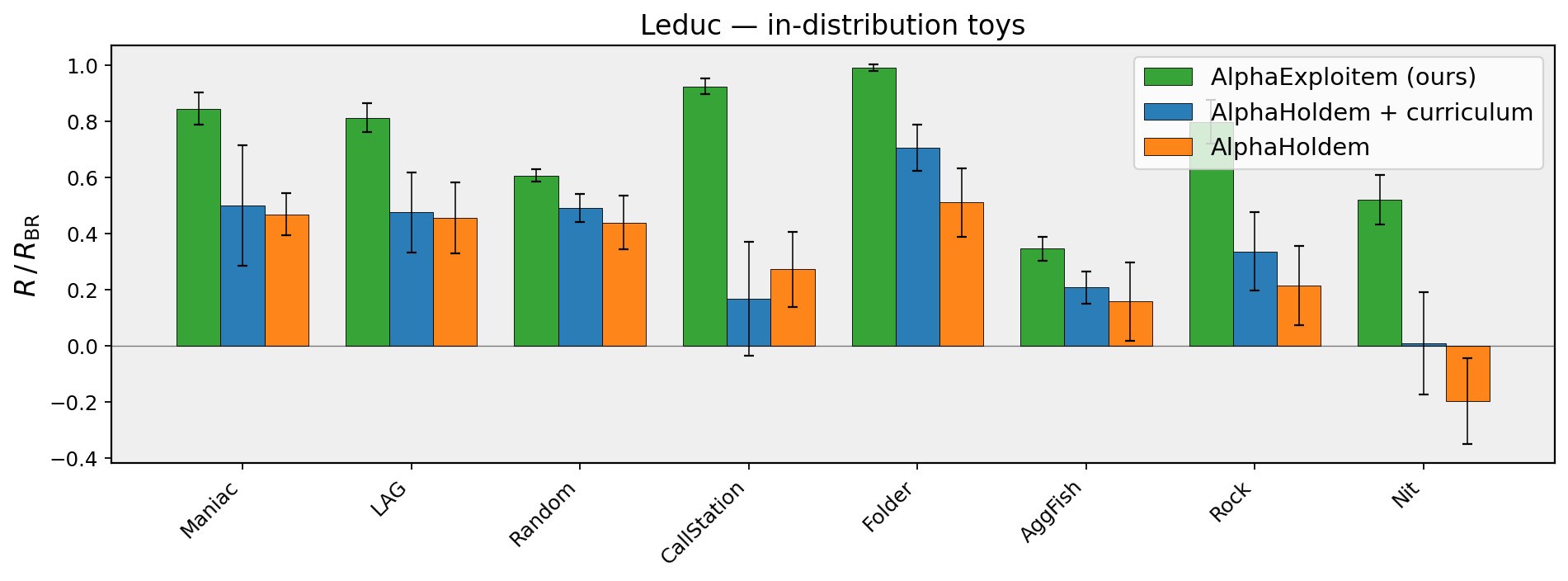}
\caption{Leduc --- in-distribution toys. Same conventions as Figure~\ref{fig:per-toy-brfrac-kuhn-id}}
\label{fig:per-toy-brfrac-leduc-id}
\end{figure}

\begin{figure}[H]
\centering
\includegraphics[width=0.92\linewidth]{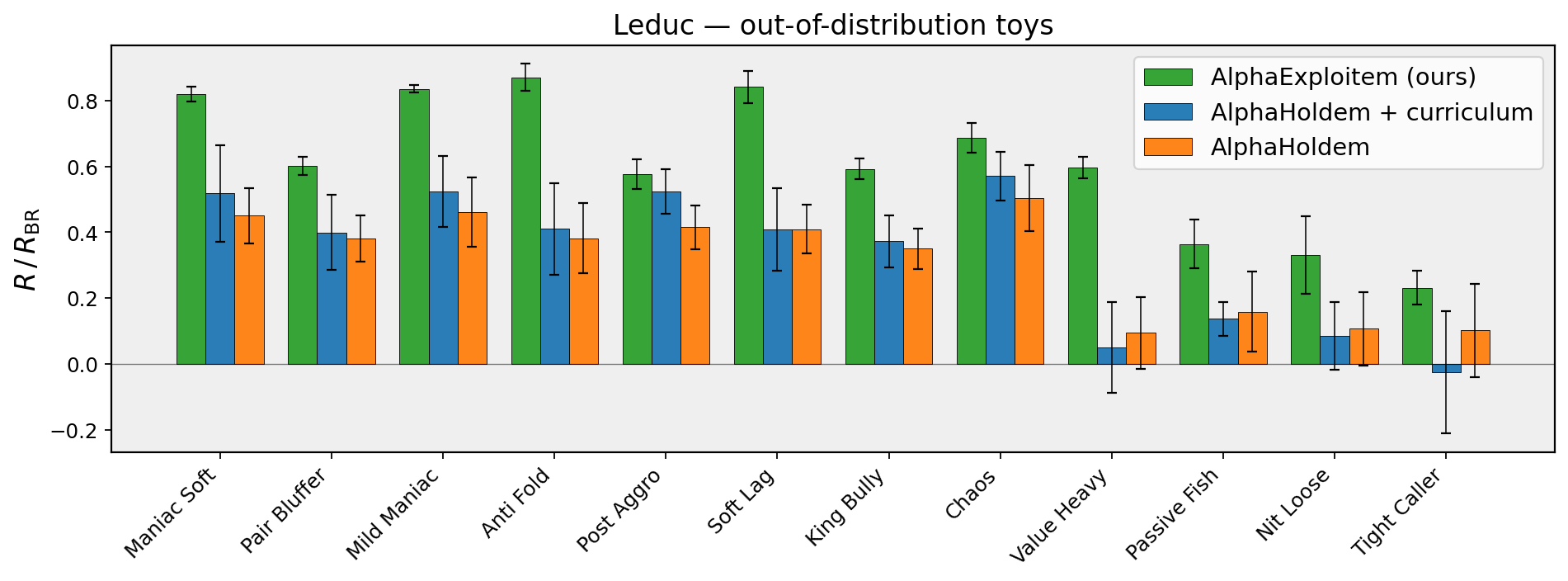}
\caption{Leduc --- out-of-distribution toys. Same conventions as Figure~\ref{fig:per-toy-brfrac-kuhn-id}}
\label{fig:per-toy-brfrac-leduc-ood}
\end{figure}

\paragraph{BR-normalised per-opponent results.} Figures~\ref{fig:per-toy-brfrac-kuhn-id}, \ref{fig:per-toy-brfrac-kuhn-ood}, \ref{fig:per-toy-brfrac-leduc-id}, \ref{fig:per-toy-brfrac-leduc-ood} restate the per-toy bars of Section~\ref{sec:results-per-toy} in BR-fraction units $R / R_\text{BR}(\pi^o)$. This makes opponents with very different exploit potential directly comparable: a value of $1.0$ means saturating the oracle ceiling, $0.5$ means realising half of the available exploit, and $0$ means baseline play. The main agent saturates a noticeably larger fraction of the ceiling than either no-encoder group on every comparable opponent; the gap to the league-only baseline is largest, in BR-fraction terms, on the moderately-exploitable opponents where there is the most room for context-conditional play to pay off.

\begin{figure}[H]
\centering
\includegraphics[width=\linewidth]{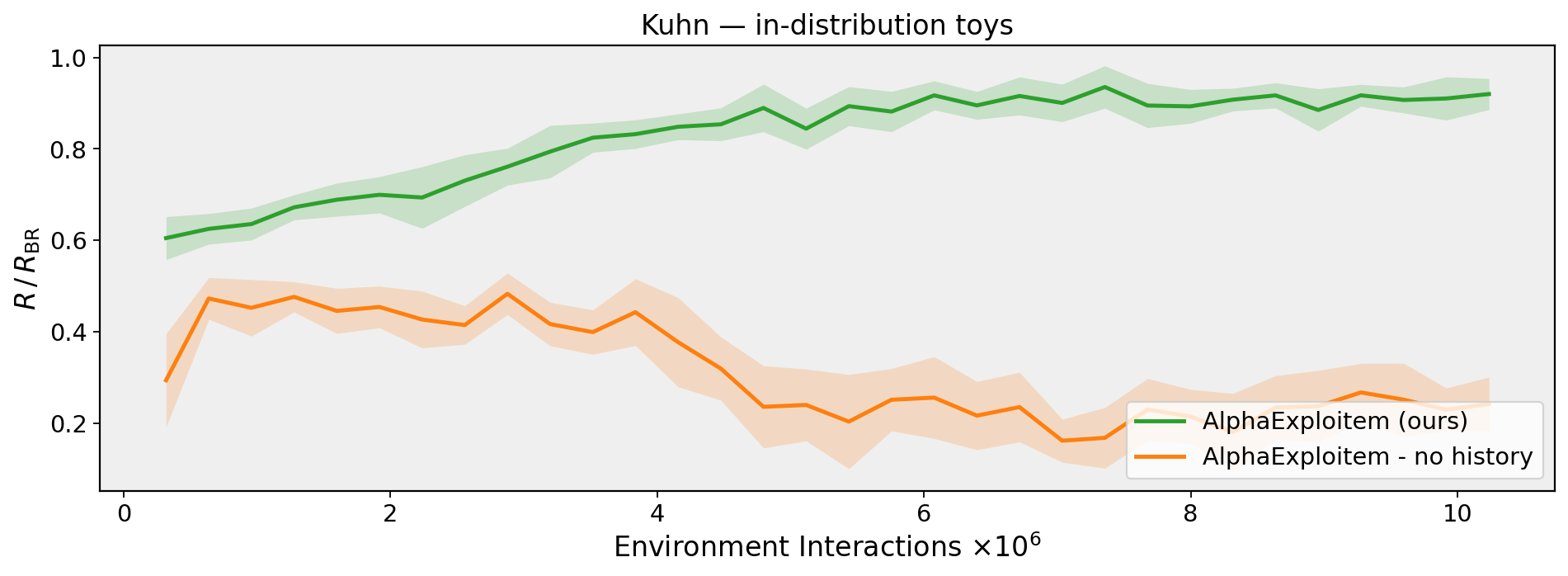}
\caption{Kuhn in-distribution BR-fraction evolution. AlphaExploitem (green) and AlphaExploitem with the cross-hand context masked at evaluation (orange). Per-epoch seed-mean of the toy-averaged $R/R_\text{BR}$ with a 95\% confidence shaded band. N = 8 seeds.}
\label{fig:brfrac-evolution-kuhn-id}
\end{figure}

\begin{figure}[H]
\centering
\includegraphics[width=\linewidth]{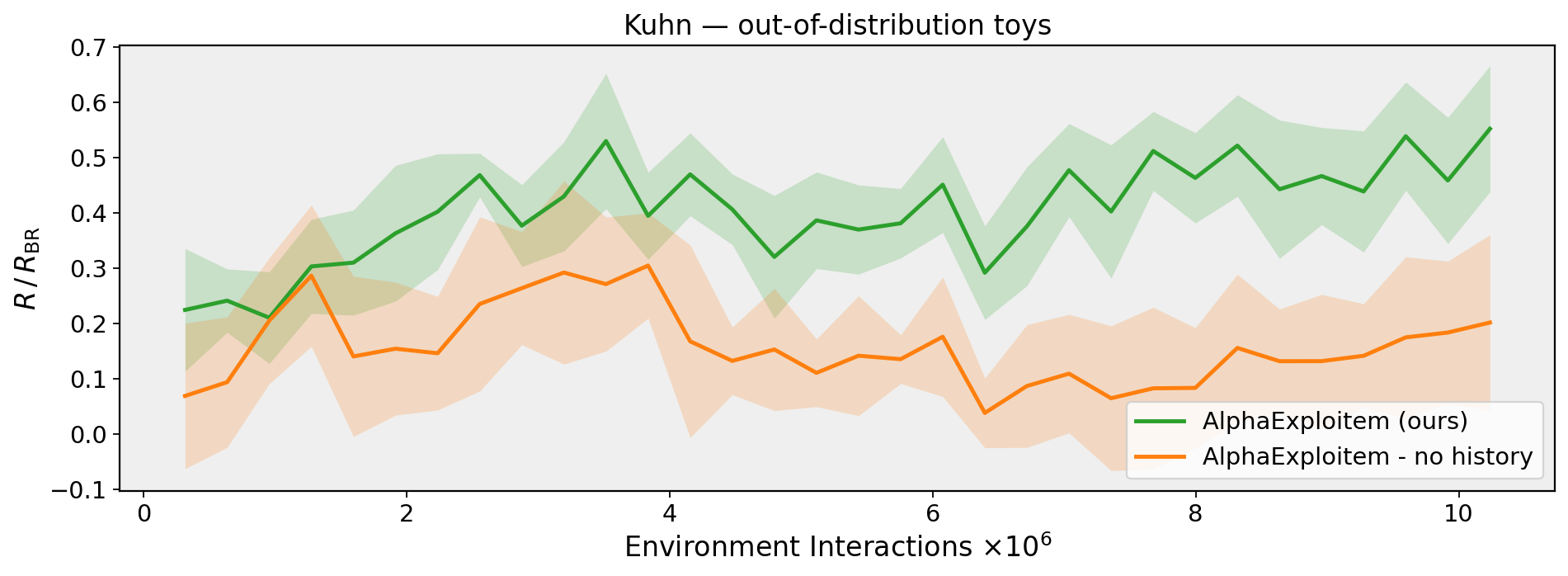}
\caption{Kuhn --- out-of-distribution. Same conventions as Figure~\ref{fig:brfrac-evolution-kuhn-id}}
\label{fig:brfrac-evolution-kuhn-ood}
\end{figure}

\begin{figure}[H]
\centering
\includegraphics[width=\linewidth]{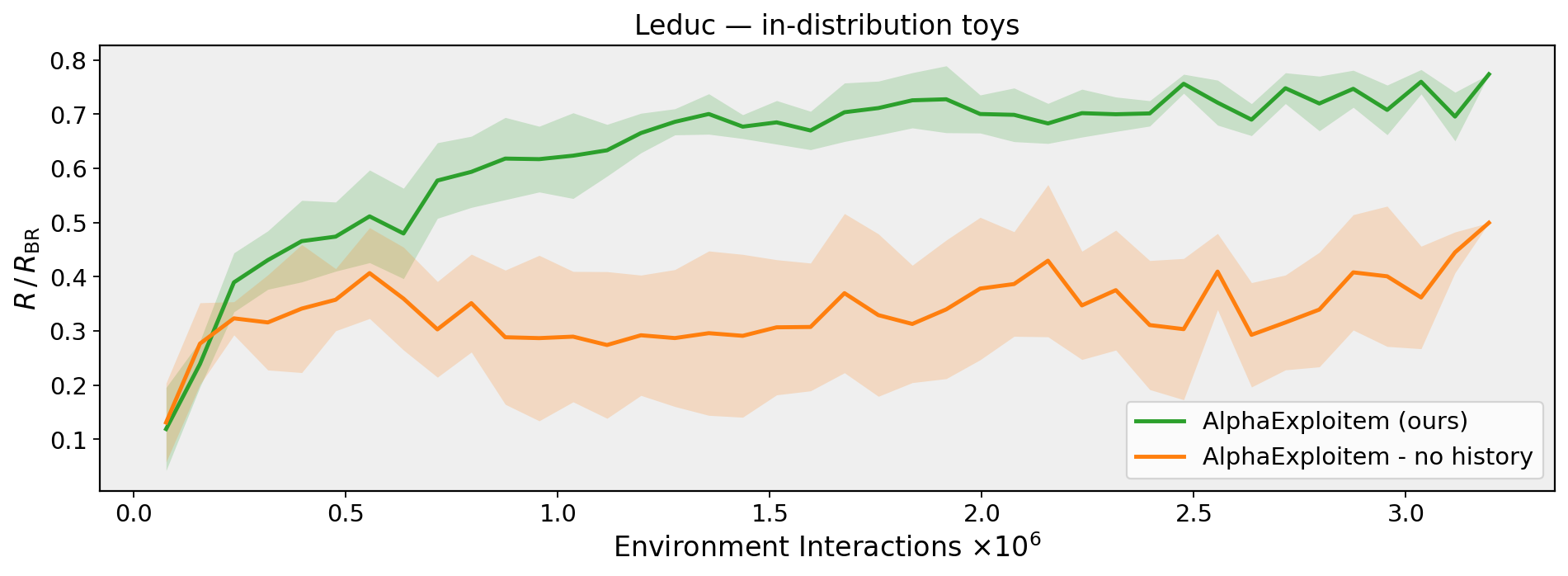}
\caption{Leduc --- in-distribution. Same conventions as Figure~\ref{fig:brfrac-evolution-kuhn-id}}
\label{fig:brfrac-evolution-leduc-id}
\end{figure}

\begin{figure}[H]
\centering
\includegraphics[width=\linewidth]{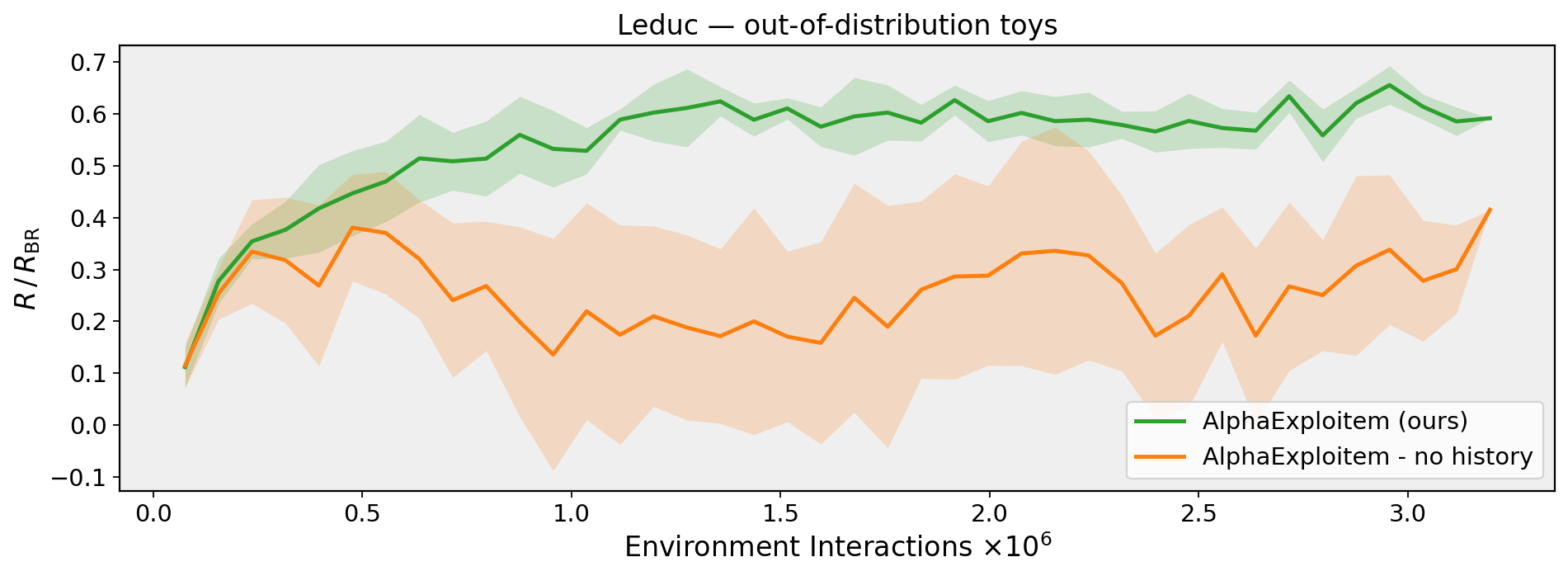}
\caption{Leduc --- out-of-distribution. Same conventions as Figure~\ref{fig:brfrac-evolution-kuhn-id}}
\label{fig:brfrac-evolution-leduc-ood}
\end{figure}

\paragraph{BR-fraction evolution.} The bars above are read at the final checkpoint, but the BR-fraction also varies smoothly over training. Figures~\ref{fig:brfrac-evolution-kuhn-id}, \ref{fig:brfrac-evolution-kuhn-ood}, \ref{fig:brfrac-evolution-leduc-id}, \ref{fig:brfrac-evolution-leduc-ood} report the per-epoch toy-averaged $R / R_\text{BR}$ for AlphaExploitem on each game-pool combination, alongside the same trained policy with the cross-hand context masked at evaluation. \texttt{ood\_p2} (Pro2 / Nash on Kuhn) is omitted from the toy-average for the same near-zero-ceiling reason given for the histograms.

%% file: Sections/Appendices/Hyperparameters.tex
\section{Hyperparameters}
\label{app:hyperparams}

\subsection{Hyperparameter tuning}
\label{app:hyperparams-tuning}

The training pipeline inherits most hyperparameters from AlphaHoldem~\citep{AlphaHoldem}: K-best league self-play, the actor-critic backbone, and the AdamW + KL-based early-stopping schedule. The PPO objective is the standard clipped surrogate~\citep{PPO}.
The novel choices in this work --- the hierarchical history transformer, the toy curriculum, and the snapshot buffer --- introduced a small number of additional hyperparameters whose values we set as follows.
For \textbf{Kuhn}, we ran a four-cell sweep over batch size, learning rate, and episode length and selected the cell with the best out-of-distribution reward and a complete training budget within the wall-clock cap, see Section~\ref{sec:results-evolution} for the comparison.
For \textbf{Leduc}, the league-only baseline collapses under the main agent's hyperparameters and was retuned via a small entropy / learning-rate / train-steps sweep. The main agent and the curriculum-matched ablation share a single hyperparameter set.

\subsection{Network architecture}
\label{app:hyperparams-arch}

The architecture and its three input streams are described in Section~\ref{sec:method} and depicted in Figure~\ref{fig:arch}.
The card and current-hand action streams are processed by lightweight per-stream linear encoders.
The cross-hand history stream is processed by a hierarchical transformer (within-hand encoder followed by an across-hand encoder), and the three streams are fused by a 2-layer MLP that splits into a softmax policy head and a scalar value head.
For the no-encoder ablation groups in both games we set \texttt{encoder\_type=none}, which short-circuits the cross-hand stream while keeping the rest of the architecture identical.

\subsection{Hyperparameters list}
\label{app:hyperparams-list}

Tables~\ref{tab:hps_kuhn} and~\ref{tab:hps_leduc} list the hyperparameters used for the main agent on Kuhn Poker and Leduc Hold'em respectively. We use $8$ seeds per experiment. 

\begin{table}[H]
\centering
\caption{Hyperparameters used for Kuhn Poker training.}
\label{tab:hps_kuhn}
\begin{tabular}{@{}llc@{}}
\toprule
\textbf{Category} & \textbf{Hyperparameter} & \textbf{Value} \\ \midrule
Architecture
 & encoder layers                      & $1$ \\
 & token embedding dim                 & $8$ \\
 & attention heads                     & $4$ \\
 & FFN width                           & $64$ \\
 & dropout prob                        & $0.1$ \\
 & cards-out / actions-out width       & $8$ / $8$ \\
 & MLP hidden width / layers           & $128$ / $2$ \\
 & policy head                         & $2$-way softmax \\
 & dtype / param dtype                 & bf16 / fp32 \\ \midrule
Trajectory generation
 & envs per opponent                   & $4$ \\
 & league size                         & $8$ \\
 & episode length                      & $100$ \\
PPO / optimizer
 & optimizer                           & AdamW \\
 & learning rate                       & $1 \times 10^{-4}$ \\
 & entropy coeff                       & $0.025$ \\
 & value-loss coeff                    & $0.01$ \\
 & PPO clip $\epsilon$                 & $0.1$ \\
 & gradient clipping                   & global-norm $1.0$ \\
 & train steps per epoch               & $5$ \\
 & train batch size                    & $64$ \\
 & minibatches per epoch               & $5$ \\
 & checkpoint frequency (epochs)       & $100$ \\ \bottomrule
\end{tabular}
\end{table}

\begin{table}[H]
\centering
\caption{Hyperparameters used for Leduc Hold'em training.}
\label{tab:hps_leduc}
\begin{tabular}{@{}llc@{}}
\toprule
\textbf{Category} & \textbf{Hyperparameter} & \textbf{Value} \\ \midrule
Architecture
 & encoder layers                      & $1$ \\
 & token embedding dim                 & $16$ \\
 & attention heads                     & $2$ \\
 & FFN width                           & $1024$ \\
 & dropout prob                        & $0.1$ \\
 & cards-out / actions-out width       & $64$ / $128$ \\
 & MLP hidden width / layers           & $128$ / $2$ \\
 & policy head                         & $3$-way softmax \\
 & dtype / param dtype                 & bf16 / fp32 \\ \midrule
Trajectory generation
 & envs per opponent                   & $8$ \\
 & league size                         & $5$ \\
 & long-tail buffer size               & $25$ \\
 & episode length                      & $100$ \\ \midrule
PPO / optimizer
 & optimizer                           & AdamW \\
 & learning rate                       & $1 \times 10^{-4}$ \\
 & entropy coeff                       & $0.005$--$0.01$ \\
 & value-loss coeff                    & $0.01$ \\
 & PPO clip $\epsilon$                 & $0.1$ \\
 & gradient clipping                   & global-norm $1.0$ \\
 & train steps per epoch               & $10$ \\
 & train batch size                    & $8$ \\
 & minibatches per epoch               & $4$ \\
 & checkpoint frequency (epochs)       & $20$ \\ \bottomrule
\end{tabular}
\end{table}

%% file: checklist.tex
\section*{NeurIPS Paper Checklist}

\begin{enumerate}

\item {\bf Claims}
    \item[] Question: Do the main claims made in the abstract and introduction accurately reflect the paper's contributions and scope?
    \item[] Answer: \answerYes{}
    \item[] Justification: We prove that the agent actively uses the game history to deviate from it's baseline strategy since the performance against weak opponents improves when the transformer history is not masked. Concurrently, AlphaExploitem is able to discern strong policies and play robustly against them, proven by its performance against the NE policy and the other strong toy policies.

\item {\bf Limitations}
    \item[] Question: Does the paper discuss the limitations of the work performed by the authors?
    \item[] Answer: \answerYes{}
    \item[] Justification: We present 2 limitations: finite time horizon and game scale. The assumptions made were stationary evaluation policies, heads-up play (one opponent) and infinite money. As for computational efficiency, the algorithm is highly parallelizable, but data inefficient.

\item {\bf Theory assumptions and proofs}
    \item[] Question: For each theoretical result, does the paper provide the full set of assumptions and a complete (and correct) proof?
    \item[] Answer: \answerNA{}
    \item[] Justification: No theoretical results are presented in the paper.

    \item {\bf Experimental result reproducibility}
    \item[] Question: Does the paper fully disclose all the information needed to reproduce the main experimental results of the paper to the extent that it affects the main claims and/or conclusions of the paper (regardless of whether the code and data are provided or not)?
    \item[] Answer: \answerYes{}
    \item[] Justification: Section \ref{sec:method} and Appendix \ref{app:hyperparams} include the necessary implementation details and hyperparameters required to reproduce the results.

\item {\bf Open access to data and code}
    \item[] Question: Does the paper provide open access to the data and code, with sufficient instructions to faithfully reproduce the main experimental results, as described in supplemental material?
    \item[] Answer: \answerNo{}
    \item[] Justification: The code will be publicly released.

\item {\bf Experimental setting/details}
    \item[] Question: Does the paper specify all the training and test details (e.g., data splits, hyperparameters, how they were chosen, type of optimizer) necessary to understand the results?
    \item[] Answer: \answerYes{}
    \item[] Justification: The experimental settings are provided in Section~\ref{sec:method} and in Appendix~\ref{app:hyperparams}.

\item {\bf Experiment statistical significance}
    \item[] Question: Does the paper report error bars suitably and correctly defined or other appropriate information about the statistical significance of the experiments?
    \item[] Answer: \answerYes{}
    \item[] Justification: All the plots include statistical significance visualizations.

\item {\bf Experiments compute resources}
    \item[] Question: For each experiment, does the paper provide sufficient information on the computer resources (type of compute workers, memory, time of execution) needed to reproduce the experiments?
    \item[] Answer: \answerYes{}
    \item[] Justification: The experimental details are presented in Section~\ref{sec:method} and in Appendix~\ref{app:hyperparams}.
    
\item {\bf Code of ethics}
    \item[] Question: Does the research conducted in the paper conform, in every respect, with the NeurIPS Code of Ethics \url{https://neurips.cc/public/EthicsGuidelines}?
    \item[] Answer: \answerYes{}
    \item[] Justification: The research conforms in every respect with the NeurIPS Code of Ethics.

\item {\bf Broader impacts}
    \item[] Question: Does the paper discuss both potential positive societal impacts and negative societal impacts of the work performed?
    \item[] Answer: \answerNA{}
    \item[] Justification: \TODO{No significant risk, }
    
\item {\bf Safeguards}
    \item[] Question: Does the paper describe safeguards that have been put in place for responsible release of data or models that have a high risk for misuse (e.g., pre-trained language models, image generators, or scraped datasets)?
    \item[] Answer: \answerNA{}
    \item[] Justification: The paper does not pose such risks.

\item {\bf Licenses for existing assets}
    \item[] Question: Are the creators or original owners of assets (e.g., code, data, models), used in the paper, properly credited and are the license and terms of use explicitly mentioned and properly respected?
    \item[] Answer: \answerNA{}
    \item[] Justification: No used assets.

\item {\bf New assets}
    \item[] Question: Are new assets introduced in the paper well documented and is the documentation provided alongside the assets?
    \item[] Answer: \answerNA{}
    \item[] Justification: No new assets are released.

\item {\bf Crowdsourcing and research with human subjects}
    \item[] Question: For crowdsourcing experiments and research with human subjects, does the paper include the full text of instructions given to participants and screenshots, if applicable, as well as details about compensation (if any)? 
    \item[] Answer: \answerNA{}
    \item[] Justification: The paper does not involve crowdsourcing or research with human subjects.

\item {\bf Institutional review board (IRB) approvals or equivalent for research with human subjects}
    \item[] Question: Does the paper describe potential risks incurred by study participants, whether such risks were disclosed to the subjects, and whether Institutional Review Board (IRB) approvals (or an equivalent approval/review based on the requirements of your country or institution) were obtained?
    \item[] Answer: \answerNA{}
    \item[] Justification: The paper does not involve crowdsourcing nor research with human subjects.

\item {\bf Declaration of LLM usage}
    \item[] Question: Does the paper describe the usage of LLMs if it is an important, original, or non-standard component of the core methods in this research? Note that if the LLM is used only for writing, editing, or formatting purposes and does \emph{not} impact the core methodology, scientific rigor, or originality of the research, declaration is not required.
    \item[] Answer: \answerNA{}
    \item[] Justification: The core method development in this research does not involve LLMs.

\end{enumerate}

%% file: references.bib
@article{DeepStack,
  author = {Moravc{\'i}k, Matej and Schmid, Martin and Burch, Neil and Lis{\'y}, Viliam and Morrill, Dustin and Bard, Nolan and Davis, Trevor and Waugh, Kevin and Johanson, Michael and Bowling, Michael},
  title = {{DeepStack}: Expert-level artificial intelligence in heads-up no-limit poker},
  journal = {Science},
  volume = {356},
  number = {6337},
  pages = {508--513},
  year = {2017}
}

@article{Pluribus,
  author = {Brown, Noam and Sandholm, Tuomas},
  title = {Superhuman {AI} for multiplayer poker},
  journal = {Science},
  volume = {365},
  number = {6456},
  pages = {885--890},
  year = {2019}
}

@article{Libratus,
  author = {Brown, Noam and Sandholm, Tuomas},
  title = {Superhuman {AI} for heads-up no-limit poker: {Libratus} beats top professionals},
  journal = {Science},
  volume = {359},
  number = {6374},
  pages = {418--424},
  year = {2018}
}

@inproceedings{AlphaHoldem,
  title = {{AlphaHoldem}: High-performance artificial intelligence for heads-up no-limit poker via end-to-end reinforcement learning},
  author = {Zhao, Enmin and Yan, Renye and Li, Jinqiu and Li, Kai and Xing, Junliang},
  booktitle = {Proceedings of the AAAI Conference on Artificial Intelligence},
  volume = {36},
  number = {4},
  pages = {4689--4697},
  year = {2022}
}

@article{Transformers,
  author = {Vaswani, Ashish and Shazeer, Noam and Parmar, Niki and Uszkoreit, Jakob and Jones, Llion and Gomez, Aidan N. and Kaiser, Lukasz and Polosukhin, Illia},
  title = {Attention is All You Need},
  journal = {CoRR},
  volume = {abs/1706.03762},
  year = {2017}
}

@article{NashEq,
  author = {Nash, John F.},
  title = {Equilibrium points in $n$-person games},
  journal = {Proceedings of the National Academy of Sciences},
  volume = {36},
  number = {1},
  pages = {48--49},
  year = {1950}
}

@article{Exploitability,
  title = {Note on non-cooperative convex games},
  journal = {Pacific Journal of Mathematics},
  author = {Nikaid{\^o}, Hukukane and Isoda, Kazuo},
  year = {1955}
}

@article{Leduc,
  title = {Bayes' bluff: Opponent modelling in poker},
  author = {Southey, Finnegan and Bowling, Michael P. and Larson, Bryce and Piccione, Carmelo and Burch, Neil and Billings, Darse and Rayner, Chris},
  journal = {arXiv preprint arXiv:1207.1411},
  year = {2012}
}

@article{kuhn2016simplified,
  title = {A simplified two-person poker},
  author = {Kuhn, Harold W.},
  journal = {Contributions to the Theory of Games},
  volume = {1},
  pages = {97--103},
  year = {1950}
}

@article{PPO,
  author = {Schulman, John and Wolski, Filip and Dhariwal, Prafulla and Radford, Alec and Klimov, Oleg},
  title = {Proximal Policy Optimization Algorithms},
  journal = {CoRR},
  volume = {abs/1707.06347},
  year = {2017}
}

@misc{jax2018github,
  author = {Bradbury, James and Frostig, Roy and Hawkins, Peter and Johnson, Matthew James and Leary, Chris and Maclaurin, Dougal and Necula, George and Paszke, Adam and Vander{P}las, Jake and Wanderman-{M}ilne, Skye and Zhang, Qiao},
  title = {{JAX}: composable transformations of {P}ython+{N}um{P}y programs},
  url = {http://github.com/jax-ml/jax},
  year = {2018}
}

@article{Loki,
  title = {Opponent modeling in poker},
  author = {Billings, Darse and Papp, Denis and Schaeffer, Jonathan and Szafron, Duane},
  journal = {AAAI/IAAI},
  volume = {493},
  number = {499},
  pages = {105},
  year = {1998}
}

@article{Poki,
  title = {The challenge of poker},
  author = {Billings, Darse and Davidson, Aaron and Schaeffer, Jonathan and Szafron, Duane},
  journal = {Artificial Intelligence},
  volume = {134},
  number = {1},
  pages = {201--240},
  year = {2002}
}

@inproceedings{KuhnExploiter,
  title = {Effective short-term opponent exploitation in simplified poker},
  author = {Hoehn, Bret and Southey, Finnegan and Holte, Robert C. and Bulitko, Valeriy},
  booktitle = {AAAI},
  volume = {5},
  pages = {783--788},
  year = {2005}
}

@article{Explicit-opponent-model,
  author = {Xu, Jiahui and Chen, Jing and Chen, Shaofei},
  title = {Efficient Opponent Exploitation in No-Limit {Texas Hold'em} Poker: A Neuroevolutionary Method Combined with Reinforcement Learning},
  journal = {Electronics},
  volume = {10},
  number = {17},
  year = {2021}
}

@misc{L2E,
  title = {{L2E}: Learning to Exploit Your Opponent},
  author = {Wu, Zhe and Li, Kai and Zhao, Enmin and Xu, Hang and Zhang, Meng and Fu, Haobo and An, Bo and Xing, Junliang},
  year = {2021},
  eprint = {2102.09381},
  archivePrefix = {arXiv}
}

@article{pokergpt,
  title = {{PokerGPT}: An End-to-End Lightweight Solver for Multi-Player {Texas Hold'em} via Large Language Model},
  author = {Huang, Chenghao and Cao, Yanbo and Wen, Yinlong and Zhou, Tao and Zhang, Yanru},
  journal = {arXiv preprint arXiv:2401.06781},
  year = {2024}
}

@inproceedings{lstm-exploiter1,
  title = {Evolving adaptive {LSTM} poker players for effective opponent exploitation},
  author = {Li, Xun and Miikkulainen, Risto},
  booktitle = {AAAI Workshops},
  year = {2017}
}

@article{lstm-exploiter2,
title={Dynamic Adaptation and Opponent Exploitation in Computer Poker},
author={Xun Li and Risto Miikkulainen},
booktitle={AAAI-18 Workshop for Imperfect Information Games},
address={New Orleans, LA, USA},
url="http://www.cs.utexas.edu/users/ai-lab?li:aaaiws2018",
year={2018}
}

@inproceedings{rnn-exploiter,
  title = {Opponent modeling and exploitation in poker using evolved recurrent neural networks},
  author = {Li, Xun and Miikkulainen, Risto},
  booktitle = {Proceedings of the Genetic and Evolutionary Computation Conference},
  pages = {189--196},
  year = {2018}
}

@article{HumansExploit,
author = {Frey, Seth and Albino, Dominic K. and Williams, Paul L.},
title = {Synergistic Information Processing Encrypts Strategic Reasoning in Poker},
journal = {Cognitive Science},
volume = {42},
number = {5},
pages = {1457-1476},
doi = {https://doi.org/10.1111/cogs.12632},
url = {https://onlinelibrary.wiley.com/doi/abs/10.1111/cogs.12632},
eprint = {https://onlinelibrary.wiley.com/doi/pdf/10.1111/cogs.12632},
year = {2018}
}

@misc{ICE,
      title={In-Context Exploiter for Extensive-Form Games}, 
      author={Shuxin Li and Chang Yang and Youzhi Zhang and Pengdeng Li and Xinrun Wang and Xiao Huang and Hau Chan and Bo An},
      year={2024},
      eprint={2408.05575},
      archivePrefix={arXiv},
      primaryClass={cs.AI},
      url={https://arxiv.org/abs/2408.05575}, 
}

@misc{pgx,
      title={Pgx: Hardware-Accelerated Parallel Game Simulators for Reinforcement Learning}, 
      author={Sotetsu Koyamada and Shinri Okano and Soichiro Nishimori and Yu Murata and Keigo Habara and Haruka Kita and Shin Ishii},
      year={2024},
      eprint={2303.17503},
      archivePrefix={arXiv},
      primaryClass={cs.AI},
      url={https://arxiv.org/abs/2303.17503}, 
}

@article{ComputerPoker,
title = {Computer poker: A review},
journal = {Artificial Intelligence},
volume = {175},
number = {5},
pages = {958-987},
year = {2011},
note = {Special Review Issue},
issn = {0004-3702},
doi = {https://doi.org/10.1016/j.artint.2010.12.005},
url = {https://www.sciencedirect.com/science/article/pii/S0004370211000191},
author = {Jonathan Rubin and Ian Watson}
}

@article{NFSP,
  title={Deep reinforcement learning from self-play in imperfect-information games},
  author={Heinrich, Johannes and Silver, David},
  journal={arXiv preprint arXiv:1603.01121},
  year={2016}
}

@InProceedings{DeepCFR,
  title = 	 {Deep Counterfactual Regret Minimization},
  author =       {Brown, Noam and Lerer, Adam and Gross, Sam and Sandholm, Tuomas},
  booktitle = 	 {Proceedings of the 36th International Conference on Machine Learning},
  pages = 	 {793--802},
  year = 	 {2019},
  editor = 	 {Chaudhuri, Kamalika and Salakhutdinov, Ruslan},
  volume = 	 {97},
  series = 	 {Proceedings of Machine Learning Research},
  month = 	 {09--15 Jun},
  publisher =    {PMLR},
  url = 	 {https://proceedings.mlr.press/v97/brown19b.html}
}

@article{DREAM,
  title={Dream: Deep regret minimization with advantage baselines and model-free learning},
  author={Steinberger, Eric and Lerer, Adam and Brown, Noam},
  journal={arXiv preprint arXiv:2006.10410},
  year={2020}
}

@article{ReBeL,
  title={Combining deep reinforcement learning and search for imperfect-information games},
  author={Brown, Noam and Bakhtin, Anton and Lerer, Adam and Gong, Qucheng},
  journal={Advances in neural information processing systems},
  volume={33},
  pages={17057--17069},
  year={2020}
}

@misc{AlgorithmDistillation,
      title={In-context Reinforcement Learning with Algorithm Distillation}, 
      author={Michael Laskin and Luyu Wang and Junhyuk Oh and Emilio Parisotto and Stephen Spencer and Richie Steigerwald and DJ Strouse and Steven Hansen and Angelos Filos and Ethan Brooks and Maxime Gazeau and Himanshu Sahni and Satinder Singh and Volodymyr Mnih},
      year={2022},
      eprint={2210.14215},
      archivePrefix={arXiv},
      primaryClass={cs.LG},
      url={https://arxiv.org/abs/2210.14215}, 
}

@misc{LLMBenchmark,
      title={GTO Wizard Benchmark},
      author={Marc-Antoine Provost and Nejc Ilenic and Christopher Solinas and Philippe Beardsell},
      year={2026},
      eprint={2603.23660},
      archivePrefix={arXiv},
      primaryClass={cs.AI},
      url={https://arxiv.org/abs/2603.23660},
}

@inproceedings{Elo1978TheRO,
  title={The rating of chessplayers, past and present},
  author={Arpad E. Elo},
  year={1978},
  url={https://api.semanticscholar.org/CorpusID:142610973}
}
